\documentclass[11pt]{article}

\usepackage[preprint]{acl}

\usepackage{times}
\usepackage{latexsym}

\usepackage[T1]{fontenc}

\usepackage[utf8]{inputenc}

\usepackage{microtype}

\usepackage{inconsolata}

\usepackage{graphicx}

\usepackage{hyperref}       
\usepackage{url}            
\usepackage{booktabs}       
\usepackage{amsfonts}       
\usepackage{nicefrac}       
\usepackage{xcolor}         
\usepackage[most]{tcolorbox}
\usepackage{cleveref} 
\usepackage{multirow}
\usepackage{placeins}
\usepackage{colortbl}
\usepackage{siunitx}
\usepackage{xspace}
\usepackage{caption}
\usepackage{bm}
\usepackage{algorithm}
\usepackage{algpseudocode}
\usepackage{rotating}
\usepackage{fontawesome5}
\usepackage{enumitem}

\usepackage[most]{tcolorbox}
\usepackage{xcolor}

\usepackage{pifont}
\newcommand{\cmark}{\textcolor{green!50!black}{\ding{51}}}
\newcommand{\xmark}{\textcolor{red!70!black}{\ding{55}}}

\usepackage{fontawesome5} 

\newtcolorbox[auto counter, number within=section]{promptbox}[2][]{
    enhanced,
    breakable,
    colback=blue!2,
    colframe=blue!50!black,
    boxrule=0.6pt,
    arc=1.5mm,
    left=1.5mm,
    right=1.5mm,
    top=2mm,
    bottom=1mm,
    before skip=6pt,
    after skip=6pt,
    fonttitle=\small\bfseries,
    coltitle=black,
    title={\faKeyboard\hspace{0.45em}Prompt~\thetcbcounter: #2},
    attach boxed title to top left={xshift=1.5mm,yshift=-1.7mm},
    boxed title style={
        colback=blue!10,
        colframe=blue!50!black,
        boxrule=0.5pt,
        arc=1mm,
        left=1mm,
        right=1mm,
        top=0.4mm,
        bottom=0.4mm
    },
    #1
}

\newcommand{\name}[1]{\textsc{SeKV}\xspace}

\title{\name{}: Resolution-Adaptive KV Cache with Hierarchical Semantic Memory for Long-Context LLM Inference}

\author{
\textbf{
Amirhossein Abaskohi$^{1}$\thanks{Corresponding author: \texttt{aabaskoh@cs.ubc.ca}. Part of this work was done during the internship at Microsoft Research~(MSR), Vancouver Lab.}\,\,\,
Giuseppe Carenini$^{1}$\,\,\,
Peter West$^{1}$\,\,\,
Yuhang He$^{2}$
}\\[0.5em]
$^{1}$University of British Columbia\,\,\,\,\, $^{2}$Microsoft Research
}

\begin{document}
\maketitle

\begin{abstract}
    Large language models increasingly operate over long contexts, where the KV cache grows linearly with sequence length and dominates memory during decoding. Token eviction discards entries permanently, and semantic compression fixes each unit's resolution at prefill, so detail that becomes relevant later cannot be recovered. We propose \textbf{\name{}}, a resolution-adaptive KV cache built around a trained zoom-in mechanism that decides at decode time which compressed spans to restore to token resolution. \name{} segments the context at high-surprisal positions, keeps anchor tokens and span summaries on GPU, and offloads low-rank factors of each span's keys and values to CPU. Lightweight per-head routers select query-relevant spans, which are reconstructed with a learned per-span rank. The cache is lossy but non-evicting: every span stays in attention, coarse or reconstructed. The base LLM stays frozen, and the router adds under 0.1\% parameters. Across five backbones and four benchmarks at a 10\% GPU-resident KV budget, \name{} outperforms every method that fixes resolution at prefill, improving over the strongest semantic baseline by 2.9 points (5.5\% relative). It matches exact-recall offloading caches on LongBench and InfiniteBench and trails on synthetic exact retrieval, while transferring 3$\times$ less data per decoding step than ShadowKV and using the least total memory among the methods we profile. At 128K context, it halves peak GPU memory and cuts total memory by one third relative to full KV caching.\footnote{Code is available on \href{https://github.com/AmirAbaskohi/SeKV}{{\textcolor{black}{\faGithub}}~GitHub.}}.
\end{abstract}

\section{Introduction}
\label{sec:introduction}

\begin{figure*}[t]
    \centering
    \includegraphics[width=0.95\textwidth]{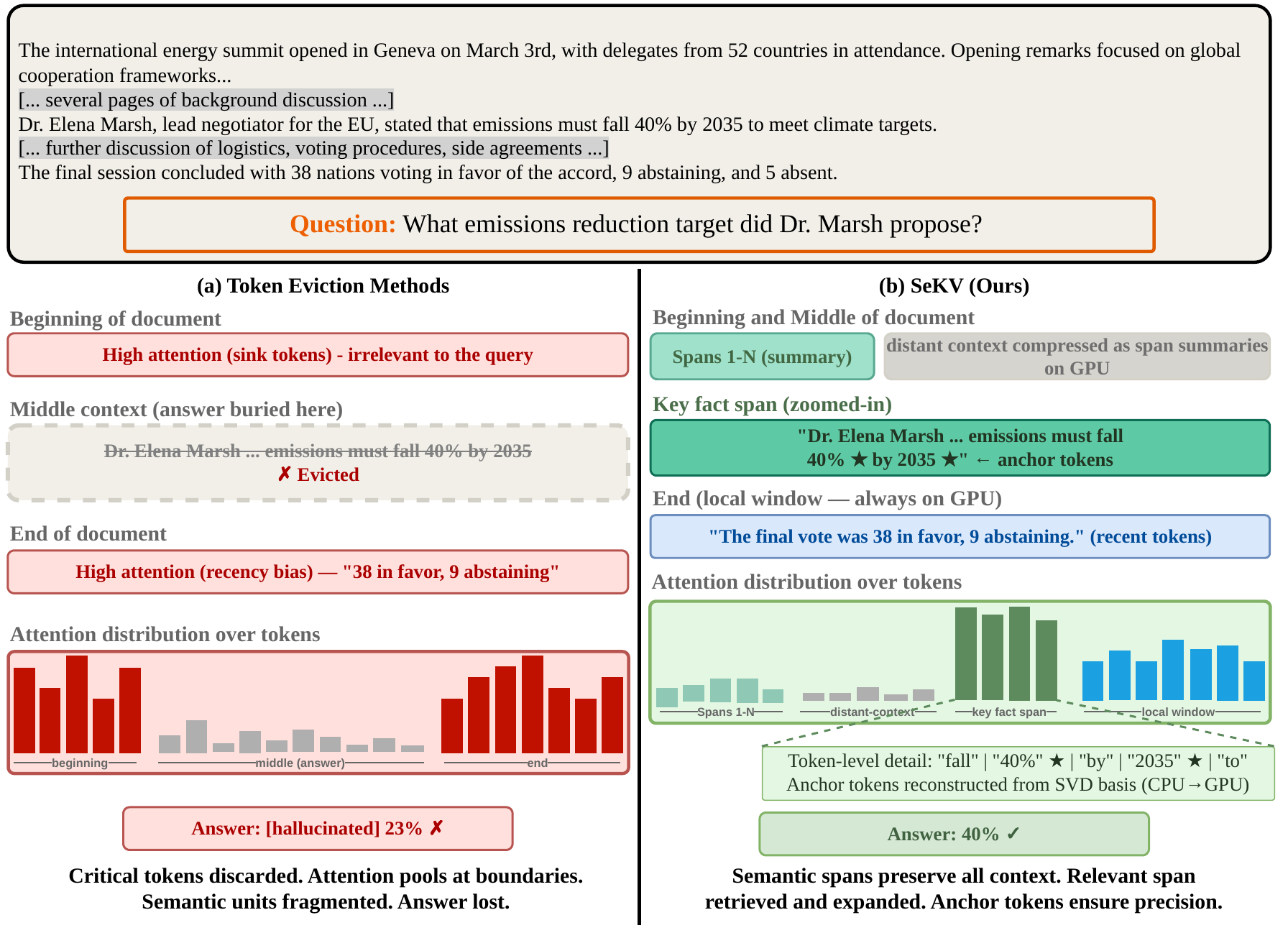}
    \caption{\textbf{(a)} Token eviction discards distant evidence, so attention pools at document boundaries and the model hallucinates. \textbf{(b)} \name{} keeps every span in attention at coarse resolution and uses a trained zoom-in mechanism to restore the query-relevant span to token resolution.}
    \label{fig:teaser}
    \vspace{-1.5em}
\end{figure*}

Large language models~(LLMs) are increasingly used over long inputs in applications such as document analysis~\cite{10656083}, multi-turn dialogue~\cite{zhang2025surveymultiturninteractioncapabilities}, and agentic planning~\cite{huang2024understandingplanningllmagents}. Although recent models accept 128K tokens or more, the Key-Value (KV) cache grows linearly with sequence length and must be kept throughout decoding, so long-context inference is bounded by GPU memory rather than by the context window itself.

KV cache compression methods trade fidelity for memory in different ways. Eviction methods keep a subset of tokens chosen by attention statistics, layer budgets, or learned retention scores, and discard the rest permanently~\cite{li2024snapkv, cai2025pyramidkv, kim2025kvzip, ICLR2026_c3887741}. Semantic methods group tokens into chunks, clusters, or sentences, but fix the resolution of each unit at prefill time~\cite{liu2026chunkkv, wu2026semanticacheefficientkvcache, zhu2025sentencekv}. Retrieval and offloading methods keep entries recoverable and recall them by query~\cite{10.5555/3737916.3741717, pmlr-v235-tang24l, chen2024arkvale, sun2025shadowkv}, but they pay for exact recall: they keep exact keys or values in CPU memory and copy the selected entries to GPU at every decoding step, or they rely on proxies whose quality degrades on context-intensive tasks~\cite{bocharnikov2026kvcacheoffloadingcontextintensive}. In long-context understanding, evidence that looks peripheral during prefill often becomes decisive for a later query, but it rarely needs to be restored exactly. We therefore argue that the cache should store context at variable resolution and decide, per query and per head, which parts to restore to token-level detail and at what fidelity.

We introduce \textbf{\name{}} (\textbf{\underline{Se}}mantic \textbf{\underline{KV}} Cache), a resolution-adaptive KV cache built around a \textbf{trained zoom-in mechanism} (Figure~\ref{fig:teaser}). Following prior work that uses surprisal to locate semantic boundaries~\cite{zhao2024meta, zhong2026semanticchunkingentropynatural}, \name{} splits the context at high-surprisal positions and keeps these \textbf{anchor tokens} at full resolution on GPU. Each span between anchors keeps a compact \textbf{summary} on GPU and \textbf{low-rank factors} of its keys and values on CPU. At each decoding step, lightweight per-head routers score spans against the current query, and the selected spans are reconstructed with a learned per-span rank and attended at token resolution. The cache is \textbf{lossy but non-evicting}: reconstruction is approximate, but every span remains in attention, so evidence ignored early can still be recovered later. The base LLM stays frozen, and the router adds under 0.1\% parameters. Across five backbones and four long-context benchmarks at a matched 10\% GPU-resident KV budget, \name{} outperforms every method that fixes resolution at prefill, improving over the strongest semantic baseline by 2.9 points (5.5\% relative). Against offloading caches that recall exact KV, it matches their accuracy on LongBench and InfiniteBench and trails on synthetic exact retrieval, while moving 3$\times$ less data per decoding step than ShadowKV and using the least total memory among the methods we profile. At 128K context, it reduces peak GPU memory by about half and total memory by one third relative to full KV caching.

In summary, our contributions are: \textbf{(1)} \name{}, a lossy but non-evicting KV cache that keeps every span at coarse resolution on GPU and at low-rank token resolution on CPU, building on surprisal-based segmentation and low-rank KV compression from prior work~\cite{zhao2024meta, saxena-etal-2024-eigen, zhu-etal-2026-ojakv}. \textbf{(2)} A \textbf{trained zoom-in mechanism}, the core of \name{}, which decides per query, layer, and head which spans to reconstruct and at what rank, trained on a frozen backbone; replacing it with training-free routing costs 2.8--3.8 points. \textbf{(3)} We evaluate \name{} against 14 baselines spanning eviction, semantic, low-rank, and offloading methods, with full GPU, CPU, and transfer accounting. \name{} is the strongest cache that does not keep exact KV, is competitive with exact-recall caches on realistic long-context tasks at a fraction of their transfer and memory cost, and narrows its gap to them on exact retrieval as the budget grows. We also analyze reconstruction quality, router behavior, and multi-turn and many-shot settings.
\section{Related Work}
\label{sec:related_work}

\paragraph{Compression at a fixed resolution.}
Most KV cache methods decide once how much of each token to keep. Eviction methods discard entries based on attention statistics, recency, or layer and entropy budgets~\cite{zhang2023ho, ICLR2024_5e5fd18f, li2024snapkv, cai2025pyramidkv, wan-etal-2025-meda, behnam2025rocketkv, kim2025kvzip}, and Locret and TRIM-KV learn retention scores on a frozen backbone~\cite{huang2025locret, ICLR2026_c3887741}. Head-aware methods assign each head a static retrieval or streaming role offline~\cite{xiao2025duoattention, ICLR2025_2a98af4f}. Low-rank methods project keys and values onto a shared or calibrated basis of uniform rank~\cite{saxena-etal-2024-eigen, zhang2024lorclowrankcompressionllms, pmlr-v300-lesens26a, chang2025xkvcrosslayersvdkvcache, zhu-etal-2026-ojakv}, and semantic methods group tokens into chunks, clusters, or sentences, some using surprisal to place boundaries~\cite{zhao2024meta, zhong2026semanticchunkingentropynatural, liu2026chunkkv, wu2026semanticacheefficientkvcache, zhu2025sentencekv}. In all of these, the resolution at which each part of the context is kept is fixed once compression is applied, so detail that becomes relevant later cannot be restored. \name{} reuses two of these ideas, surprisal-based segmentation and low-rank factorization, but as a storage format rather than a final compression decision. Its learned component also plays a different role: Locret and TRIM-KV learn what to discard, whereas \name{} learns what to restore, and unlike static head roles, its per-head expansion depends on the query and varies across tasks (Appendix~\ref{app:additional_zoom_in}).

\paragraph{Recallable caches.}
A second line keeps inactive context recoverable. Quest~\cite{pmlr-v235-tang24l} selects pages by query-aware bounds while keeping the full cache on GPU. InfLLM, ArkVale, ClusterKV, and RetroInfer offload blocks, pages, clusters, or indexed tokens to CPU and recall them through representative keys or vector search~\cite{10.5555/3737916.3741717, chen2024arkvale, 10.1109/DAC63849.2025.11132479, chen2025retroinfervectorstorageapproachscalable, zhao2026unifyingsparseattentionhierarchical}. ShadowKV~\cite{sun2025shadowkv} keeps low-rank keys on GPU, offloads exact values, and reconstructs selected tokens, and trainable sparse attention learns selection inside the model at the cost of adapting the backbone~\cite{zhao2025infllmv2densesparseswitchableattention}. These methods recall context at a single resolution, exact or through a fixed proxy, so they either store the full cache somewhere or rely on proxies that can fail on context-intensive tasks~\cite{bocharnikov2026kvcacheoffloadingcontextintensive}. \name{} differs from both lines in what it decides at decode time: every span stays in attention at coarse resolution, and a router trained on a frozen backbone chooses, per query, layer, and head, which spans to restore and at what learned rank. Appendix~\ref{app:positioning} summarizes these differences.
\section{\name{}}
\label{sec:method}

\begin{figure*}[t]
\centering
\includegraphics[width=\textwidth]{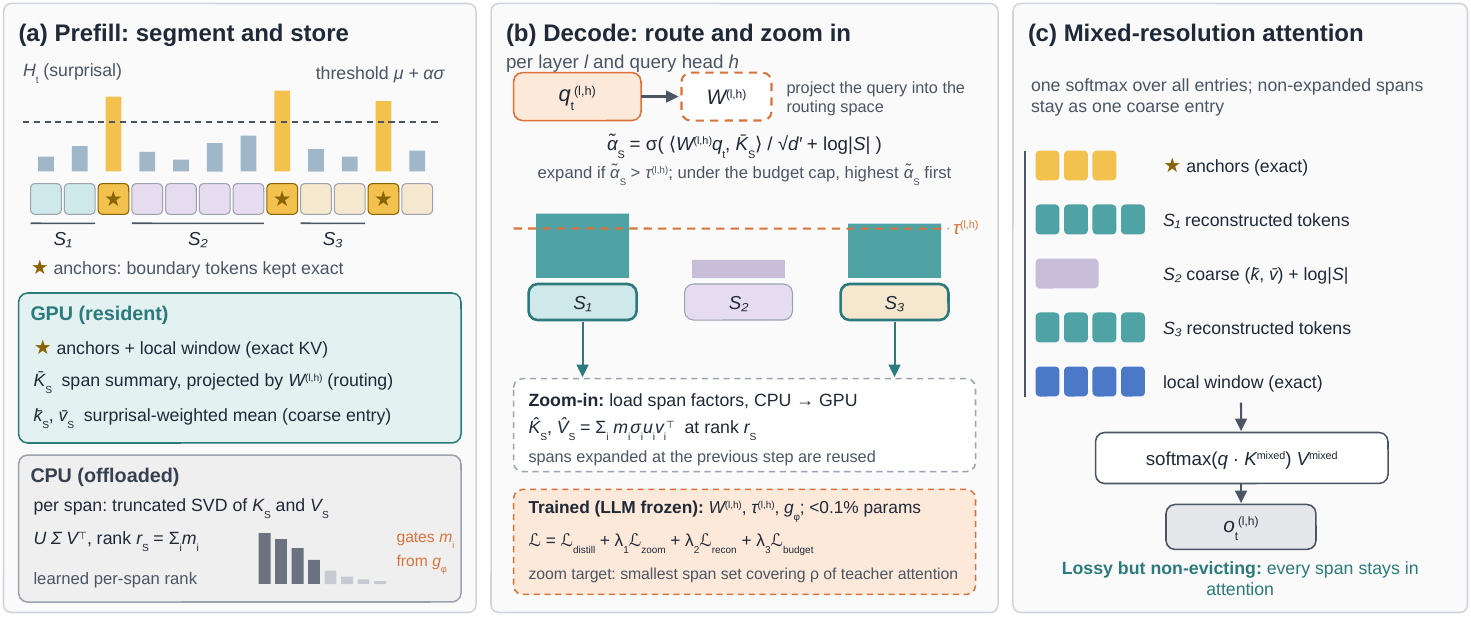}
\caption{\textbf{Overview of \name{}.} (a) Prefill splits the context at high-surprisal anchors; each span keeps a routing summary and a coarse mean on GPU and low-rank key and value factors with a learned rank on CPU. (b) At each decoding step, a trained per-head gate selects spans to expand, subject to the GPU budget. (c) Expanded spans enter attention at token resolution, the rest as one coarse entry, in a single softmax.}
\label{fig:method}
\vspace{-1.5em}
\end{figure*}

\name{} stores every span of the context at two resolutions and decides at decode time which spans to restore to token level (Figure~\ref{fig:method}). Section~\ref{sec:memory} describes what is stored where, Section~\ref{sec:zoom} the trained zoom-in mechanism, and Section~\ref{sec:training} how it is trained. We omit layer and head indices $(l,h)$ where the meaning is clear.

\subsection{Span Memory}
\label{sec:memory}

\paragraph{Segmentation.}
Following prior work that uses surprisal to locate semantic boundaries~\cite{zhao2024meta, zhong2026semanticchunkingentropynatural}, we compute $H_t = -\log p(x_t \mid x_{<t})$ during prefill and mark positions with $H_t > \mu + \alpha\sigma$ as boundaries, where $\mu$ and $\sigma$ are the sequence mean and standard deviation, and boundaries closer than $L_{\text{gap}}$ tokens are merged, keeping the highest-surprisal one. In practice, anchors make up about 1\% of tokens and spans average 113 to 131 tokens across benchmarks and backbones (Table~\ref{tab:e6}). Boundary tokens become \textbf{anchors}, kept as exact KV on GPU, and the tokens between consecutive anchors form a span $S$. Surprisal requires the LM head at every prompt position, which we apply in chunks; this adds 5--6\% to prefill time (Appendix~\ref{app:efficiency}).

\paragraph{GPU: routing summary and coarse entry.}
Each span keeps two small GPU-resident representations per head. The \emph{routing summary} projects token keys into a $d'$-dimensional space with a learned per-head matrix $W \in \mathbb{R}^{d' \times d_h}$ and averages them with surprisal weights $w_t = H_t / \sum_{k \in S} H_k$:
\begin{equation}
\bar{K}_S = \frac{\sum_{t \in S} w_t W K_t}{\big\|\sum_{t \in S} w_t W K_t\big\|_2}.
\label{eq:summary_key}
\end{equation}
The \emph{coarse entry} is the surprisal-weighted mean key and value in the original space, $\bar{k}_S = \sum_{t} w_t K_t$ and $\bar{v}_S = \sum_{t} w_t V_t$, which represents the span in attention when it is not expanded.

\paragraph{CPU: low-rank factors with a learned rank.}
For each span and head, we factorize the pre-RoPE key matrix $K_S \in \mathbb{R}^{|S| \times d_h}$ and the value matrix $V_S$ with a truncated SVD up to rank $R_{\max}$, and a lightweight predictor $g_\phi$ assigns each component a gate $m_i \in (0,1)$ from the singular spectrum and the span summary:
\begin{equation}
\hat{X}_S = \sum_{i=1}^{R_{\max}} m_i\, \sigma_i\, \mathbf{a}_i \mathbf{b}_i^{\top}, \qquad X \in \{K_S, V_S\},
\label{eq:reconstruction}
\end{equation}
where $\sigma_i$, $\mathbf{a}_i$, and $\mathbf{b}_i$ are the singular values and vectors. Components with negligible gates are dropped, so the stored rank $r_S$ adapts per span. Factors are stored in fp32 on CPU and cast to bf16 after transfer. RoPE is applied to reconstructed keys at their original positions. Spans shorter than $L_{\min}$ stay exact on GPU.

\subsection{Trained Zoom-In}
\label{sec:zoom}

At each decoding step, each head scores every span with an independent gate rather than a softmax, since expansion is a per-span decision:
\begin{equation}
\begin{aligned}
\tilde{\alpha}_S
&= \mathrm{sig}\!\left(
\frac{(W q_t)^{\top} \bar{K}_S}{\sqrt{d'}}
+ \log |S|
\right), \\
z_S
&= \mathbf{1}\big[\tilde{\alpha}_S > \tau\big],
\end{aligned}
\label{eq:routing_gate}
\end{equation}
where $\tau$ is a learned per-head threshold and $\log|S|$ is a size prior that keeps long spans from being under-selected. Per-head thresholds let heads that retrieve distant evidence expand often while others rarely do. Routing is per query head, while reconstruction is per key-value head under grouped-query attention: a span is reconstructed for a key-value head if any query head in its group selects it, but only the selecting heads attend to it at token resolution, and the others keep its coarse entry.

\paragraph{Budget.} We define the budget as the peak GPU-resident KV during a decoding step: anchors, local window, summaries, coarse entries, and all reconstructed spans. When the spans with $z_S = 1$ exceed it, we keep them in decreasing order of $\tilde{\alpha}_S$ until it is filled, so $\tau$ sets the operating point and the cap enforces the matched budget.

\paragraph{Mixed-resolution attention.} Each head then attends once over a single memory: anchors and the local window as exact KV, expanded spans as reconstructed tokens (Eq.~\ref{eq:reconstruction}), and every other span as its coarse entry $(\bar{k}_S, \bar{v}_S)$ with $\log|S|$ added to its score so that one entry competes in proportion to the tokens it stands for~\cite{NEURIPS2023_ab05dc8b}. Because all entries share one softmax, zooming in refines a span's contribution, and no span leaves attention: the cache is lossy but non-evicting. Factors are transferred asynchronously, overlapping with computation, and spans expanded at the previous step stay on GPU and are reused, which avoids most transfers (Section~\ref{sec:efficiency_analysis}).

\subsection{Training}
\label{sec:training}

The base LLM stays frozen. We train only the projections $W$, the thresholds $\tau$, and the rank predictor $g_\phi$, under 0.1\% of the backbone's parameters, once per backbone on generic text (Section~\ref{sec:exp_setup}). The objective combines four terms:
\begin{equation}
\mathcal{L} = \mathcal{L}_{\text{distill}} + \lambda_1 \mathcal{L}_{\text{zoom}} + \lambda_2 \mathcal{L}_{\text{recon}} + \lambda_3 \mathcal{L}_{\text{budget}}.
\label{eq:total_loss}
\end{equation}

\paragraph{Distillation.} Output distributions are aligned with a full-KV teacher:
\begin{equation}
\mathcal{L}_{\text{distill}} = D_{\text{KL}}\big(p_{\text{full}} \,\|\, p_{\name{}}\big).
\label{eq:loss_distill}
\end{equation}
Gradients reach $W$ and $\tau$ through a straight-through estimator for $z_S$~\cite{bengio2013estimatingpropagatinggradientsstochastic}.

\paragraph{Zoom supervision.} Let $a_S = \sum_{t \in S} \alpha_t^{\text{teacher}}$ be the teacher's attention mass on span $S$, and $\mathcal{C}_\rho$ the smallest set of spans whose cumulative mass reaches $\rho$. With target $y_S = \mathbf{1}[S \in \mathcal{C}_\rho]$,
\begin{equation}
\mathcal{L}_{\text{zoom}} = \sum_{l,h,S} \mathrm{BCE}_{w^+}\big(\tilde{\alpha}_S,\, y_S\big),
\label{eq:loss_zoom}
\end{equation}
where $w^+$ up-weights the rare relevant spans. The target depends only on the teacher, so the threshold cannot define its own supervision.

\paragraph{Reconstruction.} Because singular components are orthonormal, the gated error of a factorization is $\sum_i (1-m_i)^2 \sigma_i^2$. Summing over keys and values, heads, layers, and spans,
\begin{equation}
\mathcal{L}_{\text{recon}} = \sum_{l,h,S} \sum_{X \in \{K_S, V_S\}} \sum_i \big(1 - m_i^{X}\big)^2 \big(\sigma_i^{X}\big)^2,
\label{eq:loss_recon}
\end{equation}
which trains $g_\phi$ while treating the SVD outputs as constants, avoiding unstable gradients through the decomposition.

\paragraph{Budget.} Without a cost term, the objectives above favor expanding every span at full rank:
\begin{equation}
\mathcal{L}_{\text{budget}} = \sum_{l,h,S} \Big(z_S + \beta \sum_i m_i\Big).
\label{eq:loss_budget}
\end{equation}
The first term pushes thresholds up when expansions are costly, and $\beta$ sets the cost of retained rank relative to expanded spans.
\section{Experiments and Results}
\label{sec:experiments}

\begin{table*}[t]
\centering
\small
\setlength{\tabcolsep}{4.2pt}
\resizebox{\textwidth}{!}{%
\begin{tabular}{l cccc cccc ccccc}
\toprule
& \multicolumn{4}{c}{\textbf{Llama-3.1-8B-Instruct}} & \multicolumn{4}{c}{\textbf{Qwen2.5-14B-Instruct}} & \multicolumn{5}{c}{\textbf{Average over five backbones}} \\
\cmidrule(lr){2-5}\cmidrule(lr){6-9}\cmidrule(lr){10-14}
\textbf{Method} & LB & RULER & IB & NIAH & LB & RULER & IB & NIAH & LB & RULER & IB & NIAH & \textbf{All} \\
\midrule
Full KV & 50.42 & 87.91 & 45.13 & 97.11 & 54.68 & 89.28 & 48.71 & 98.22 & 47.59 & 77.57 & 37.32 & 89.03 & 62.88 \\
\midrule
\rowcolor{gray!12}\multicolumn{14}{l}{\emph{Eviction}} \\
StreamingLLM & 36.25 & 51.24 & 22.93 & 54.07 & 40.90 & 46.71 & 25.27 & 55.73 & 34.18 & 40.46 & 18.37 & 46.69 & 34.92 \\
H2O & 41.31 & 60.67 & 30.69 & 67.97 & 44.01 & 59.62 & 32.50 & 68.26 & 37.93 & 50.51 & 24.23 & 58.25 & 42.73 \\
SnapKV & 45.27 & 70.75 & 34.94 & 82.69 & 47.48 & 69.72 & 37.50 & 84.23 & 41.47 & 59.17 & 28.27 & 72.67 & 50.40 \\
PyramidKV & 45.07 & 71.48 & 35.03 & 85.92 & 48.66 & 71.18 & 38.50 & 84.51 & 41.37 & 60.26 & 28.17 & 73.96 & 50.94 \\
RocketKV & 45.47 & 78.66 & 38.52 & 90.06 & 49.91 & 78.03 & 42.50 & 89.86 & 42.21 & 65.83 & 31.50 & 77.44 & 54.24 \\
\rowcolor{gray!12}\multicolumn{14}{l}{\emph{Learned eviction}} \\
TRIM-KV & 45.86 & 76.09 & 36.94 & 84.57 & 49.73 & 73.59 & 39.94 & 85.56 & 41.93 & 62.14 & 29.73 & 73.50 & 51.83 \\
\rowcolor{gray!12}\multicolumn{14}{l}{\emph{Semantic compression}} \\
ChunkKV & 45.45 & 74.88 & 36.52 & 83.52 & 49.39 & 72.18 & 40.31 & 81.21 & 41.93 & 61.71 & 29.63 & 71.59 & 51.21 \\
SemantiCache & 45.17 & 73.32 & 36.27 & 82.68 & 49.24 & 73.40 & 40.60 & 82.82 & 41.96 & 61.15 & 29.82 & 71.88 & 51.20 \\
SentenceKV & 46.06 & 73.70 & 36.12 & 82.54 & 49.30 & 73.83 & 40.60 & 81.53 & 42.17 & 61.48 & 29.79 & 72.20 & 51.41 \\
\rowcolor{gray!12}\multicolumn{14}{l}{\emph{Low-rank compression}} \\
OjaKV & 38.25 & 64.18 & 31.46 & 66.53 & 43.40 & 62.83 & 32.50 & 67.97 & 36.14 & 52.67 & 25.23 & 58.69 & 43.18 \\
\rowcolor{gray!12}\multicolumn{14}{l}{\emph{Recallable caches}} \\
Quest$^\dagger$ & 46.95 & 79.92 & 39.30 & 92.12 & 50.86 & 77.64 & 42.48 & 90.23 & 43.59 & 66.03 & 31.70 & 79.64 & 55.24 \\
InfLLM & 45.23 & 72.09 & 37.29 & 90.04 & 49.61 & 67.93 & 41.04 & 89.67 & 42.12 & 58.57 & 30.25 & 78.05 & 52.25 \\
RetroInfer & 46.89 & \underline{81.26} & \textbf{39.92} & \textbf{92.86} & 51.19 & \textbf{80.26} & \underline{42.65} & \textbf{91.75} & 43.60 & \underline{68.27} & \textbf{32.12} & \textbf{80.24} & \underline{56.06} \\
ShadowKV & \underline{47.28} & \textbf{82.27} & \underline{39.49} & \underline{92.16} & \underline{51.56} & \underline{79.83} & \textbf{43.72} & \underline{91.23} & \underline{43.68} & \textbf{68.56} & \underline{32.05} & \underline{80.08} & \textbf{56.09} \\
\midrule
\rowcolor{teal!8}\textbf{\name{} (ours)} & \textbf{47.70} & 77.89 & 38.47 & 85.36 & \textbf{51.80} & 77.24 & 42.50 & 85.12 & \textbf{44.22} & 65.24 & 31.80 & 75.80 & 54.26 \\
\bottomrule
\end{tabular}%
}
\caption{Accuracy at a 10\% GPU-resident KV budget on \textsc{Llama-3.1-8B-Instruct}, \textsc{Qwen2.5-14B-Instruct}, and averaged over all five backbones (per-backbone results for all five in Appendix~\ref{app:full_results}). LB: LongBench; IB: InfiniteBench. Best compressed result in \textbf{bold}, second best \underline{underlined}. $^\dagger$Quest keeps the full KV cache on GPU and is matched on its attention-token budget. All results are means over three seeds.}
\label{tab:main_results}
\vspace{-1em}
\end{table*}

\subsection{Experimental Setup}
\label{sec:exp_setup}

\noindent
\textbf{Benchmarks.}
We evaluate on four long-context benchmarks, each with its official prompts and metrics. \textbf{LongBench}~\cite{bai-etal-2024-longbench}: the 16 English tasks, macro-averaged over task-specific metrics. \textbf{RULER}~\cite{hsieh2024ruler}: all 13 tasks at 8K, 16K, 32K, 64K, and 128K tokens, averaged over tasks and lengths. \textbf{InfiniteBench}~\cite{zhang-etal-2024-bench}: nine English, code, math, and retrieval tasks, macro-averaged. \textbf{Needle-in-a-Haystack (NIAH)}~\cite{LLMTest_NeedleInAHaystack}: a single needle at 8 lengths from 8K to 128K and 11 depths, 100 trials per cell, scored by exact substring match. Inputs longer than 128K are truncated in the middle. \name{} needs a minimum resident state on GPU (anchors, summaries, coarse entries, and the local window), so examples too short for this state to fit in the budget are excluded for all methods; at 10\% this removes about 33\% of LongBench and no examples from the other benchmarks (Appendix~\ref{app:benchmark_baseline_details}). Beyond these four benchmarks, we evaluate many-shot reasoning on GSM8K~\cite{cobbe2021trainingverifierssolvemath} with 50 in-context demonstrations, and multi-turn context reuse on SCBench~\cite{li2025scbench} with query-agnostic compression before the first turn (Appendix~\ref{app:additional_results}).

\noindent
\textbf{Baselines.}
We compare against Full KV and 14 compression methods: eviction (StreamingLLM~\cite{ICLR2024_5e5fd18f}, H2O~\cite{zhang2023ho}, SnapKV~\cite{li2024snapkv}, PyramidKV~\cite{cai2025pyramidkv}, RocketKV~\cite{behnam2025rocketkv}), learned eviction (TRIM-KV~\cite{ICLR2026_c3887741}), semantic compression (ChunkKV~\cite{liu2026chunkkv}, SemantiCache~\cite{wu2026semanticacheefficientkvcache}, SentenceKV~\cite{zhu2025sentencekv}), low-rank compression (OjaKV~\cite{zhu-etal-2026-ojakv}), and recallable caches (Quest~\cite{pmlr-v235-tang24l}, InfLLM~\cite{10.5555/3737916.3741717}, RetroInfer~\cite{chen2025retroinfervectorstorageapproachscalable}, ShadowKV~\cite{sun2025shadowkv}). Every method is configured so that its GPU-resident KV state equals the same budget, 10\% of Full KV's KV bytes unless stated otherwise; CPU memory is reported separately (Section~\ref{sec:efficiency_analysis}). Quest keeps the full cache on GPU by design, so it is matched on its attention-token budget instead. Configurations are given in Appendix~\ref{app:benchmark_baseline_details}.

\noindent
\textbf{Backbones.}
We use \textsc{Llama-3.2-3B-Instruct}, \textsc{Llama-3-8B}, \textsc{Llama-3.1-8B-Instruct}~\cite{grattafiori2024llama3herdmodels}, \textsc{Mistral-7B-Instruct-v0.3}~\cite{jiang2023mistral7b}, and \textsc{Qwen2.5-14B-Instruct}~\cite{qwen2025qwen25technicalreport}. \textsc{Llama-3-8B} and \textsc{Mistral-7B} are extended to 128K with YaRN~\cite{peng2024yarn}. All methods share the backbone, tokenizer, prompt, and greedy decoding within each setting.

\noindent
\textbf{Implementation.}
\name{} is implemented in PyTorch with HuggingFace Transformers. For each backbone, the router is trained once on long RedPajama documents~\cite{weber2024redpajama} (arXiv, books, and code), which share no data with any evaluation benchmark, for about 3K steps over 0.5B tokens; this takes 1.8 to 5.2 hours on one $8\times$A100 (80GB) node. We use $R_{\max}=32$, $d'=32$, $\alpha=1.0$, $w=512$, $L_{\min}=16$, $\lambda_1=1.0$, $\lambda_2=0.5$, $\lambda_3=0.1$, $\beta=0.1$, $w^+=10$, and $\rho=0.9$ (Appendix~\ref{app:training_details}); training cost, router transfer, and sensitivity to these values are reported in Appendix~\ref{app:hyper_sensitivity}.


\subsection{Accuracy under a Matched GPU Budget}
\label{sec:benchmark_results}

\noindent
\textbf{\name{} outperforms every method that fixes resolution at prefill.}
Table~\ref{tab:main_results} reports accuracy at a 10\% GPU-resident budget on \textsc{Llama-3.1-8B} and \textsc{Qwen2.5-14B} and averaged over all five backbones; Appendix~\ref{app:full_results} gives every backbone. Averaged over the 20 backbone and benchmark settings, \name{} improves over the strongest semantic baseline, SentenceKV, by 2.9 points (5.5\% relative), over the learned eviction method TRIM-KV by 2.4 points, and over SnapKV by 3.9 points. It outperforms SentenceKV in all 20 settings and TRIM-KV in 19. The gains over SentenceKV are largest on RULER (3.8 points) and NIAH (3.6), where specific evidence must be recovered from a long context, compared with 2.0 on LongBench and InfiniteBench. Eviction and semantic methods commit to what they keep before the query is known, so evidence they compressed or dropped cannot return; \name{} keeps every span in attention and restores the ones the query needs. OjaKV, which compresses all tokens with one low-rank basis at the same budget, trails by 11.1 points, which suggests that uniform low-rank compression at this budget loses too much detail and that rank must be spent selectively.

\noindent
\textbf{Against recallable caches, \name{} is strongest on realistic tasks and behind on exact retrieval.}
\name{} has the highest LongBench average of all compressed methods and is the best on LongBench for four of five backbones and on InfiniteBench for two (Appendix~\ref{app:full_results}). On average it trails ShadowKV and RetroInfer by 1.8 points and Quest, which keeps the full cache on GPU, by 1.0 point, and it is level with RocketKV. The deficit is concentrated on NIAH ($-4.3$ against ShadowKV) and RULER ($-3.3$), the tasks that hinge on recovering a few exact tokens, while \name{} is ahead on LongBench ($+0.5$) and level on InfiniteBench ($-0.3$). These methods recall exact keys or values from a full-precision copy of the cache, whereas \name{} reconstructs spans from low-rank factors, which is lossy for single tokens (Appendix~\ref{app:reconstruction}); Appendix~\ref{app:niah_diagnosis} traces the remaining retrieval errors to value reconstruction and to query heads that keep the coarse entry of a needle span. In exchange, \name{} stores 11--17\% less on CPU than ShadowKV and about half as much as InfLLM and SentenceKV, and has the lowest total memory among the methods we profile (Section~\ref{sec:efficiency_analysis}).

\begin{figure}[t]
    \centering
    \includegraphics[width=\linewidth]{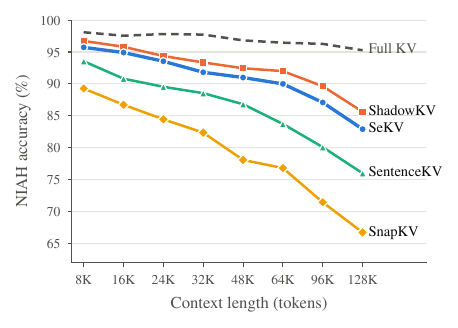}
    \caption{NIAH accuracy versus context length on \textsc{Llama-3.1-8B-Instruct} at a 10\% GPU-resident KV budget, averaged over 11 needle depths (1{,}100 trials per point). Results come from a separate diagnostic run (Appendix~\ref{app:niah_diagnosis}), so absolute values differ from Table~\ref{tab:main_results}.}
    \label{fig:niah_length}
\end{figure}

\noindent
\textbf{Robustness across backbones.}
\name{} retains 87\% of Full KV accuracy on average and 93\% on LongBench. It outperforms SentenceKV on every backbone by 2.7 to 3.0 points, and its standard deviation over three router seeds is at most 0.85 points (Appendix~\ref{app:seed_variance}), well below these margins. As for all methods, the two YaRN-extended backbones, \textsc{Llama-3-8B} and \textsc{Mistral-7B}, lose the largest share of their Full KV accuracy on RULER and NIAH. On NIAH, accuracy falls with context length for every compressed method while needle depth has little effect (Figure~\ref{fig:niah_length}; Appendix~\ref{app:niah_diagnosis}).

\subsection{Memory and Latency}
\label{sec:efficiency_analysis}

\begin{table}[t]
\centering
\small
\setlength{\tabcolsep}{3.5pt}
\resizebox{\linewidth}{!}{%
\begin{tabular}{lcccccc}
\toprule
\textbf{Method} & \textbf{GPU} & \textbf{CPU} & \textbf{Total} & \textbf{H2D} & \textbf{Decode} & \textbf{Acc.} \\
 & (GB) & (GB) & (GB) & (MB/step) & (ms/tok) & \\
\midrule
Full KV & 69.47 & 0.00 & 69.47 & 0 & 61.80 & 62.88 \\
\midrule
Quest$^\dagger$ & 71.23 & 0.00 & 71.23 & 0 & 37.89 & 55.24 \\
InfLLM & 35.06 & 24.16 & 59.22 & 151.5 & 78.37 & 52.25 \\
SentenceKV & 35.22 & 23.58 & 58.80 & 131.7 & 64.71 & 51.41 \\
ShadowKV & 35.42 & 12.84 & 48.26 & 212.8 & \textbf{31.68} & \textbf{56.09} \\
\rowcolor{teal!8}\textbf{\name{} (ours)} & 35.31 & \textbf{10.70} & \textbf{46.01} & \textbf{68.2} & 56.30 & 54.26 \\
\bottomrule
\end{tabular}%
}
\caption{Memory, transfer, and decode latency at 128K context on \textsc{Qwen2.5-14B-Instruct} (batch size 1, one A100 80GB, 10\% budget). H2D: host-to-device transfer per decoding step. Acc.: average accuracy over all backbones and benchmarks (Table~\ref{tab:main_results}). Best among compressed methods in \textbf{bold}; for CPU and H2D, among methods that offload to CPU. $^\dagger$Quest keeps the full KV cache on GPU. Results from 8K to 128K and for \textsc{Llama-3.1-8B-Instruct} are in Appendix~\ref{app:efficiency}.}
\label{tab:efficiency_128k}
\end{table}

\noindent
\textbf{Memory.}
At 128K context, \name{} reduces peak GPU memory from 69.5 GB to 35.3 GB on \textsc{Qwen2.5-14B} (49\%) and from 43.9 GB to 21.5 GB on \textsc{Llama-3.1-8B} (51\%) (Table~\ref{tab:efficiency_128k}). From 8K to 128K, its peak GPU memory grows only from 32.4 GB to 35.3 GB, while Full KV grows from 34.1 GB to 69.5 GB (Table~\ref{tab:e3a}). All offloading methods meet the same GPU budget, so they differ in what they keep on CPU. \name{} stores 10.7 GB of low-rank factors on CPU, 17\% less than ShadowKV and less than half of InfLLM and SentenceKV, which offload exact KV. Its total memory of 46.0 GB is the lowest of all methods we profile, 34\% below Full KV. Quest decodes faster than \name{}, but it keeps the full cache on GPU and saves no memory.

\noindent
\textbf{Transfer and latency.}
\name{} moves 68 MB per decoding step at 128K, 1.9--3.1$\times$ less than the other offloading methods, because only expanded spans are fetched, as low-rank factors, and 72--89\% of expanded spans are reused from the previous step (Appendix~\ref{app:efficiency}). Its decode latency is 10\% above Full KV at 8K, where routing and reconstruction outweigh the small cache savings, equal at 32K, and 9\% below at 128K (Table~\ref{tab:e3a}). \name{} decodes faster than SentenceKV and InfLLM at every length but is 1.8$\times$ slower than ShadowKV, which uses fused kernels, at 128K. Our PyTorch implementation reaches only 3--5 GB/s effective host-to-device bandwidth, and 9.6 ms of the 16.9 ms spent on transfers per step at 128K is not hidden behind computation, so pinned buffers and batched copies are a direct route to lower latency. Prefill is 13--15\% slower than Full KV, split between surprisal (5--6\%) and segmentation with SVD (7--10\%), and is paid once per context.

\subsection{What Each Component Contributes}
\label{sec:ablations}

\begin{table}[t]
\centering
\small
\setlength{\tabcolsep}{4pt}
\resizebox{\linewidth}{!}{%
\begin{tabular}{lccccc}
\toprule
& \multicolumn{2}{c}{\textbf{Qwen2.5-14B}} & \multicolumn{2}{c}{\textbf{Llama-3.1-8B}} & \textbf{CPU} \\
\cmidrule(lr){2-3}\cmidrule(lr){4-5}
\textbf{Variant} & RULER & NIAH & RULER & NIAH & (GB) \\
\midrule
\rowcolor{teal!8}\name{} (full) & 79.49 & 91.62 & 78.94 & 90.88 & 10.7 \\
\midrule
\rowcolor{gray!12}\multicolumn{6}{l}{\emph{Router}} \\
Training-free routing & $-3.2$ & $-2.8$ & $-3.8$ & $-3.4$ & 10.7 \\
Random projections & $-7.9$ & $-9.3$ & $-8.5$ & $-9.7$ & 10.7 \\
No zoom-in (coarse only) & $-6.6$ & $-8.9$ & $-7.0$ & $-9.3$ & 10.7 \\
Global threshold & $-2.1$ & $-1.9$ & $-2.4$ & $-2.6$ & 10.7 \\
\rowcolor{gray!12}\multicolumn{6}{l}{\emph{Storage}} \\
Exact KV, matched CPU bytes & $-2.9$ & $-3.5$ & $-3.4$ & $-3.2$ & 10.7 \\
Fixed rank $R=32$ & $-0.4$ & $+0.3$ & $-0.3$ & $-0.3$ & 16.8 \\
\rowcolor{gray!12}\multicolumn{6}{l}{\emph{Segmentation}} \\
Fixed-length chunks & $-2.4$ & $-2.9$ & $-2.5$ & $-2.7$ & 10.7 \\
No anchors & $-3.0$ & $-3.8$ & $-3.2$ & $-3.9$ & 10.7 \\
\bottomrule
\end{tabular}%
}
\caption{Change in accuracy (points) when one component of \name{} is removed or replaced, at a 10\% GPU-resident budget; CPU memory at 128K on \textsc{Qwen2.5-14B}. Training-free routing scores spans by $q^{\top}\bar{k}_S$ in the original key space. Fixed-length chunks match the mean span length. These runs use the diagnostic setting of Appendix~\ref{app:niah_diagnosis}, so the full-model row differs from Table~\ref{tab:main_results}.}
\label{tab:ablations}
\end{table}

Table~\ref{tab:ablations} removes or replaces one component at a time. \textbf{Router:} replacing the trained router with training-free routing in the original key space costs 2.8 to 3.8 points, and random projections or no zoom-in cost 6.6 to 9.7, so the gain comes from what the router learns, not from expanding spans alone; a single global threshold costs about 2 points, supporting per-head thresholds. \textbf{Storage:} spending the same CPU memory on exact KV for a subset of spans costs 2.9 to 3.5 points, because low-rank factors cover every span, and a fixed rank of 32 matches the learned rank in accuracy but needs 1.6$\times$ the CPU memory. \textbf{Segmentation:} fixed-length chunks of the same mean length cost 2.4 to 2.9 points and dropping anchors 3.0 to 3.9, so where spans begin and which tokens stay exact both matter.

\subsection{Sensitivity to the GPU Budget}
\label{sec:budget}

\begin{table}[t]
\centering
\small
\setlength{\tabcolsep}{3.5pt}
\resizebox{\linewidth}{!}{%
\begin{tabular}{llcccccc}
\toprule
& & \multicolumn{3}{c}{\textbf{Qwen2.5-14B}} & \multicolumn{3}{c}{\textbf{Llama-3.1-8B}} \\
\cmidrule(lr){3-5}\cmidrule(lr){6-8}
\textbf{Budget} & \textbf{Method} & RULER & NIAH & IB & RULER & NIAH & IB \\
\midrule
-- & Full KV & 85.79 & 98.12 & 48.71 & 84.23 & 97.00 & 45.13 \\
\midrule
\multirow{4}{*}{5\%} & SnapKV & 62.40 & 75.76 & 33.64 & 62.72 & 75.01 & 30.70 \\
 & SentenceKV & 69.80 & 76.85 & 38.42 & 69.24 & 77.63 & 34.38 \\
 & ShadowKV & \textbf{80.23} & \textbf{91.19} & \textbf{42.93} & \textbf{79.13} & \textbf{90.36} & \textbf{38.86} \\
 & \cellcolor{teal!8}\name{} & \cellcolor{teal!8}\underline{76.80} & \cellcolor{teal!8}\underline{88.38} & \cellcolor{teal!8}\underline{41.13} & \cellcolor{teal!8}\underline{75.82} & \cellcolor{teal!8}\underline{87.11} & \cellcolor{teal!8}\underline{36.83} \\
\midrule
\multirow{4}{*}{10\%} & SnapKV & 69.72 & 84.23 & 37.50 & 70.75 & 82.69 & 34.94 \\
 & SentenceKV & 73.83 & 81.53 & 40.60 & 73.70 & 82.54 & 36.12 \\
 & ShadowKV & \textbf{81.50} & \textbf{93.12} & \textbf{43.72} & \textbf{80.71} & \textbf{92.50} & \textbf{39.49} \\
 & \cellcolor{teal!8}\name{} & \cellcolor{teal!8}\underline{79.49} & \cellcolor{teal!8}\underline{91.62} & \cellcolor{teal!8}\underline{42.50} & \cellcolor{teal!8}\underline{78.94} & \cellcolor{teal!8}\underline{90.88} & \cellcolor{teal!8}\underline{38.47} \\
\midrule
\multirow{4}{*}{20\%} & SnapKV & 73.10 & 88.14 & 39.29 & 74.39 & 86.81 & 36.90 \\
 & SentenceKV & 76.30 & 84.26 & 41.92 & 76.38 & 84.95 & 37.30 \\
 & ShadowKV & \textbf{82.37} & \textbf{94.21} & \textbf{44.25} & \textbf{81.84} & \underline{93.42} & \textbf{39.97} \\
 & \cellcolor{teal!8}\name{} & \cellcolor{teal!8}\underline{81.41} & \cellcolor{teal!8}\underline{93.89} & \cellcolor{teal!8}\underline{43.37} & \cellcolor{teal!8}\underline{81.07} & \cellcolor{teal!8}\textbf{93.46} & \cellcolor{teal!8}\underline{39.54} \\
\bottomrule
\end{tabular}%
}
\caption{Accuracy at 5\%, 10\%, and 20\% GPU-resident KV budgets. IB: InfiniteBench. Best per budget in \textbf{bold}, second \underline{underlined}. At 5\%, all methods are scored on the examples for which \name{}'s resident state fits the budget (53--61\% of RULER and NIAH, all of InfiniteBench). Runs use the diagnostic setting of Appendix~\ref{app:niah_diagnosis}, except the 10\% SnapKV and SentenceKV rows, which come from the main runs.}
\label{tab:budget}
\end{table}

Table~\ref{tab:budget} varies the GPU-resident budget from 5\% to 20\%. All methods improve with budget, and eviction loses the most when it shrinks: from 10\% to 5\% on \textsc{Qwen2.5-14B}, SnapKV drops 7.3 points on RULER and 8.5 on NIAH, against 2.7 and 3.2 for \name{} and 1.3 and 1.9 for ShadowKV. \name{} gains more from extra budget than ShadowKV, so its gap to ShadowKV narrows from 3.4 points at 5\% to 1.0 at 20\% on RULER and from 2.8 to 0.3 on NIAH, and on \textsc{Llama-3.1-8B} \name{} matches ShadowKV on NIAH at 20\%. This is consistent with the budget cap being binding at low budgets, where relevant spans compete for expansion. Budgets of 1--2\% are below \name{}'s floor: anchors, summaries, coarse entries, and the 512-token local window need about 2.4\% of Full KV memory at 128K, and more at shorter contexts.

\subsection{Where \name{} Zooms In}
\label{sec:zoom_analysis}
 
\begin{figure}[t]
    \centering
    \includegraphics[width=\linewidth]{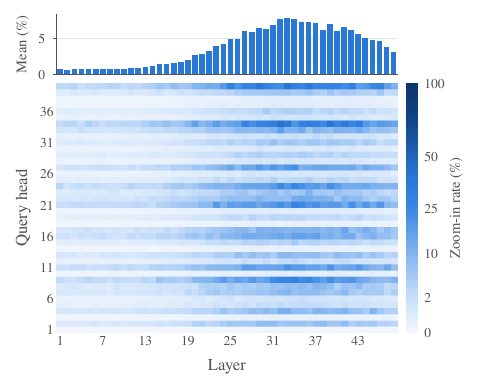}
    \caption{Zoom-in rate per layer and query head on NIAH with \textsc{Qwen2.5-14B-Instruct} at a 10\% GPU-resident KV budget. Top: mean over heads per layer. The color scale is square-root compressed.}
    \label{fig:zoom_heatmap}
\end{figure}
 
Figure~\ref{fig:zoom_heatmap} shows how often each query head expands a span on NIAH with \textsc{Qwen2.5-14B}, and Appendix~\ref{app:additional_zoom_in} extends this to three backbones and three benchmarks. Three patterns hold in all nine settings. \textbf{Expansion is sparse.} Heads expand 3.6--4.0\% of spans on NIAH, 5.2--5.7\% on RULER, and 7.4--8.1\% on InfiniteBench, so the router spends more resolution where evidence is spread over more of the context. \textbf{Expansion is concentrated.} Layers in the first third of the network expand fewer than 2\% of spans, rates peak about two thirds of the way through the network (layers 21--25 of 32 for \textsc{Llama-3.1-8B} and \textsc{Mistral-7B}, 32--36 of 48 for \textsc{Qwen2.5-14B}), and the 10\% most active heads account for 49--63\% of all expansions, while 24--52\% of heads expand fewer than 1\% of spans. \textbf{Which heads expand depends on the task.} Per-head rates for the same backbone correlate only weakly across benchmarks (Spearman 0.07--0.49), so no fixed allocation of heads chosen offline fits every task. These patterns explain two earlier results: a single global threshold cannot express this variation and costs about 2 points (Table~\ref{tab:ablations}), and because few spans are expanded and 72--89\% of them are reused from the previous step, \name{} moves only 68 MB per decoding step at 128K (Section~\ref{sec:efficiency_analysis}).
\section{Conclusion}
\label{sec:conclusion}

We introduced \name{}, a lossy but non-evicting KV cache that keeps every span in attention at coarse resolution and uses a trained zoom-in mechanism to reconstruct query-relevant spans from low-rank factors in CPU memory. At a 10\% GPU-resident budget, \name{} outperforms every method that fixes resolution at prefill and matches exact-recall offloading caches on LongBench and InfiniteBench, while moving 3$\times$ less data per decoding step than ShadowKV and halving peak GPU memory at 128K context. Closing the remaining gap on exact retrieval and integrating \name{} with fused kernels and batched serving are natural next steps.


\section*{Limitations}

\paragraph{Exact retrieval.}
\name{} reconstructs expanded spans from low-rank factors, so token-level detail is approximate. This matters most for tasks that must reproduce several isolated tokens verbatim, such as multi-key retrieval, where exact-recall offloading caches remain stronger. Our diagnosis (Appendix~\ref{app:niah_diagnosis}) shows that this gap comes from value reconstruction and routing, both of which can be improved within the same framework, for example with a higher rank for values of high-surprisal spans or with routing supervision at retrieval positions. The gap also narrows as the GPU budget grows (Table~\ref{tab:budget}).

\paragraph{Serving integration.}
Our implementation is a research prototype in PyTorch, and all efficiency results use batch size 1 on a single GPU. We do not evaluate batched or multi-request serving, and \name{} is not yet integrated with serving systems and kernels such as vLLM~\cite{10.1145/3600006.3613165} and FlashAttention~\cite{NEURIPS2022_67d57c32}. The attention over reconstructed spans is standard scaled dot-product attention, and spans map naturally to paged KV blocks, so we expect such integration, together with fused kernels and pinned transfers, to reduce latency further.

\paragraph{Training and model access.}
\name{} trains a small router once per backbone on generic text and keeps the backbone frozen. This one-time cost is a few GPU hours (Appendix~\ref{app:hyper_sensitivity}), and routers transfer within a model family with a small loss. Like other KV cache methods, \name{} requires access to model internals and therefore applies to open-weight models rather than models served only through an API.

\paragraph{Scope of evaluation.}
We evaluate English benchmarks with contexts up to 128K tokens and backbones up to 14B parameters. \name{} needs a small resident state on GPU, so very short inputs, where the KV cache is small and compression is rarely needed, fall outside its operating range (Appendix~\ref{app:benchmark_baseline_details}).

\section*{Ethics Statement}

\paragraph{Scope and inherited risks.}
\name{} is an inference-efficiency method: it changes how the KV cache of a frozen language model is stored and accessed, and it does not add new capabilities, training data, or model behaviors. It therefore inherits the risks of the underlying models, including biased, harmful, or factually incorrect outputs, and does not mitigate them.

\paragraph{Effect of lossy compression.}
Because reconstruction is lossy, \name{} can change a model's output relative to full KV caching, most often on tasks that require reproducing isolated details verbatim (Appendix~\ref{app:niah_diagnosis}). In high-stakes applications, compressed inference should be validated against the uncompressed model before deployment.

\paragraph{Licenses and intended use of artifacts.}
All artifacts we use are publicly released, and we use each only for non-commercial research, consistent with its license and intended use. The evaluation benchmarks are released under the MIT license (LongBench, GSM8K, SCBench, and the Needle-in-a-Haystack test) or the Apache 2.0 license (RULER and InfiniteBench). \textsc{Mistral-7B-Instruct-v0.3} and \textsc{Qwen2.5-14B-Instruct} are released under Apache 2.0, and the Llama backbones under the Meta Llama 3, 3.1, and 3.2 Community Licenses. The router is trained on the arXiv, books, and code subsets of RedPajama, whose subsets follow the terms of their sources.
Baselines use their official implementations under their released licenses. Our code will be released for research use, and any released router weights will follow the license of the backbone they were trained for.

\paragraph{Data and privacy.}
We collect no new data, involve no human subjects, and use all benchmarks and corpora without modification. We release no data, and the router weights cannot generate text, so our artifacts do not expose personal information or offensive content beyond what the public resources may already contain.

\paragraph{Environmental impact.}
Training the router takes 14 to 42 GPU-hours per backbone on A100 GPUs, a one-time cost that is small relative to pretraining, and the backbone is never fine-tuned. By halving peak GPU memory at long context, \name{} can reduce the hardware needed to serve long-context models.


\bibliography{custom}


\appendix

\section{Positioning Against Related Methods}
\label{app:positioning}

Table~\ref{tab:positioning} compares \name{} with the methods discussed in Section~\ref{sec:related_work} along four questions: what unit the cache operates on, how context outside the current focus is stored, whether that context can return to attention during decoding, and whether the resolution at which it returns is chosen per query. The components \name{} shares with prior work, surprisal-based segmentation and low-rank factorization, appear in other rows. The combination that no other row has is a non-evicting cache in which a router trained on a frozen backbone chooses, per query, layer, and head, which spans to restore and at what rank.

\begin{table*}[t]
\centering
\small
\resizebox{\textwidth}{!}{%
\begin{tabular}{llllcl}
\toprule
\textbf{Method} & \textbf{Unit} & \textbf{Inactive context stored as} & \textbf{Recovery during decoding} & \textbf{Per-query resolution} & \textbf{Trained} \\
\midrule
\rowcolor{gray!15}\multicolumn{6}{l}{\emph{Eviction and budget allocation}} \\
StreamingLLM~\citeyearpar{ICLR2024_5e5fd18f}, H2O~\citeyearpar{zhang2023ho} & token & discarded & none & \xmark & \xmark \\
SnapKV~\citeyearpar{li2024snapkv}, PyramidKV~\citeyearpar{cai2025pyramidkv} & token & discarded & none & \xmark & \xmark \\
MEDA~\citeyearpar{wan-etal-2025-meda} & token & discarded or merged & none & \xmark & \xmark \\
RocketKV~\citeyearpar{behnam2025rocketkv} & token & discarded after prefill & sparse selection over retained tokens & \xmark & \xmark \\
KVzip~\citeyearpar{kim2025kvzip} & token & discarded (query-agnostic) & none & \xmark & \xmark \\
Locret~\citeyearpar{huang2025locret}, TRIM-KV~\citeyearpar{ICLR2026_c3887741} & token & discarded & none & \xmark & \cmark{} retention scorer, frozen LLM \\
DuoAttention~\citeyearpar{xiao2025duoattention}, RazorAttention~\citeyearpar{ICLR2025_2a98af4f} & head & sink and recent tokens on non-retrieval heads & none & \xmark & \cmark{} per-head gates (DuoAttention) \\
\midrule
\rowcolor{gray!15}\multicolumn{6}{l}{\emph{Low-rank compression}} \\
EigenAttention~\citeyearpar{saxena-etal-2024-eigen}, LoRC~\citeyearpar{zhang2024lorclowrankcompressionllms} & token & low-rank, calibrated basis or weights (GPU) & none (always compressed) & \xmark & \xmark \\
KQ-SVD~\citeyearpar{pmlr-v300-lesens26a}, xKV~\citeyearpar{chang2025xkvcrosslayersvdkvcache} & token / layer & low-rank attention or cross-layer basis (GPU) & none (always compressed) & \xmark & \xmark \\
OjaKV~\citeyearpar{zhu-etal-2026-ojakv} & token & low-rank, online basis; anchors exact (GPU) & none (always compressed) & \xmark & \xmark \\
\midrule
\rowcolor{gray!15}\multicolumn{6}{l}{\emph{Semantic compression}} \\
ChunkKV~\citeyearpar{liu2026chunkkv} & chunk & discarded & none & \xmark & \xmark \\
SemantiCache~\citeyearpar{wu2026semanticacheefficientkvcache} & cluster & merged into representative KV & none & \xmark & \xmark \\
SentenceKV~\citeyearpar{zhu2025sentencekv} & sentence & exact KV (CPU) & sentence-level retrieval, exact & \xmark & \xmark \\
\midrule
\rowcolor{gray!15}\multicolumn{6}{l}{\emph{Recallable caches}} \\
Quest~\citeyearpar{pmlr-v235-tang24l} & page & exact KV (GPU) & query-aware page selection, exact & \xmark & \xmark \\
InfLLM~\citeyearpar{10.5555/3737916.3741717}, ArkVale~\citeyearpar{chen2024arkvale} & block / page & exact KV (CPU) & recall via representative keys or digests, exact & \xmark & \xmark \\
ClusterKV~\citeyearpar{10.1109/DAC63849.2025.11132479} & semantic cluster & exact KV & recall by cluster, exact & \xmark & \xmark \\
RetroInfer~\citeyearpar{chen2025retroinfervectorstorageapproachscalable} & token (indexed) & exact KV (CPU) & vector retrieval, exact & \xmark & \xmark \\
ShadowKV~\citeyearpar{sun2025shadowkv} & chunk & low-rank keys (GPU), exact values (CPU) & sparse reconstruction, fixed rank & \xmark & \xmark \\
InfLLM-V2~\citeyearpar{zhao2025infllmv2densesparseswitchableattention} & block & full KV (GPU) & learned sparse selection, exact & \xmark & \cmark{} backbone adapted \\
\midrule
\rowcolor{teal!10}\textbf{\name{} (ours)} & surprisal span & coarse entry (GPU); low-rank K and V (CPU) & router-selected spans reconstructed & \cmark{} span and rank & \cmark{} router, frozen LLM \\
\bottomrule
\end{tabular}%
}
\caption{Positioning of \name{} against related KV cache methods. ``Recovery during decoding'' describes whether context outside the current focus can return to attention; ``per-query resolution'' describes whether the fidelity at which it returns is decided per query rather than fixed by the method.}
\label{tab:positioning}
\end{table*}
\paragraph{Where each distinction is tested.} Each row group has an empirical counterpart in our experiments. Against learned eviction, \name{} outperforms TRIM-KV on every backbone (Table~\ref{tab:main_results}) and degrades far less across turns when context is compressed before the first query (Table~\ref{tab:e10}). Against recallable caches, \name{} trails ShadowKV by 1.8 points on average while using less CPU memory and the lowest total memory at 128K ((Tables~\ref{tab:main_results} and~\ref{tab:efficiency_128k})). The contribution of the trained router is isolated by training-free routing and random-projection controls, and the choice of low-rank storage by an exact-KV control at matched CPU bytes and a fixed-rank control (Table~\ref{tab:ablations}). Reconstruction quality of the low-rank factors is reported in Appendix~\ref{app:reconstruction}.
\section{Training Details}
\label{app:training_details}

\paragraph{Data.} \name{} trains only the routing projections, thresholds, and rank predictor (Section~\ref{sec:training}), so a single language-modeling corpus suffices. We use long documents from RedPajama~\cite{weber2024redpajama} (arXiv, books, and code subsets). The distillation and reconstruction losses use all positions; the zoom loss is supervised at sampled query positions whose teacher attention falls mostly outside the local window, and the spans they attend to form the coverage target $\mathcal{C}_\rho$. No instruction or synthetic-retrieval data is used, and the corpus shares no data with any evaluation benchmark. A separate router is trained for each backbone.

\paragraph{Optimization.} The teacher is the frozen backbone with a full KV cache. We use AdamW~\cite{loshchilov2018decoupled} with learning rate $1\times10^{-3}$, weight decay $0.01$, cosine decay, and 10\% warmup, for about 3K steps over 0.5B tokens, with a length curriculum from 8K to 32K tokens, an effective batch of 8 sequences, bf16, and early stopping on a held-out split. The budget weight $\lambda_3$ is annealed from 0 to 0.1 over the first half of training, and thresholds are initialized permissively ($\tau=0.05$) so that spans expand early and gradients flow through the straight-through estimator. Teacher span-attention targets are computed at a sampled subset of query positions to bound the cost of materializing attention weights.

\paragraph{Hyperparameters.} $\lambda_1 = 1.0$, $\lambda_2 = 0.5$, $\lambda_3 = 0.1$, $\beta = 0.1$, $w^+ = 10$, $\rho = 0.9$, $\alpha = 1.0$, $w = 512$, $L_{\min} = 16$, $d' = 32$, and $R_{\max} = 32$. Sensitivity to $R_{\max}$, $\alpha$, $w$, $L_{\min}$, and $\beta$ is reported in Appendix~\ref{app:hyper_sensitivity}.

\paragraph{Cost.} Table~\ref{tab:e9b} lists trainable parameters and training time per backbone: 2.8M to 7.9M parameters (0.054\% to 0.086\% of the backbone) and 1.8 to 5.2 hours on one $8\times$A100 (80GB) node, i.e., 14 to 42 GPU-hours.
\section{Benchmark and Baseline Details}
\label{app:benchmark_baseline_details}

\subsection{Benchmarks}

Table~\ref{tab:dataset_statistics} lists the evaluation sets. All methods are scored on the same examples. Examples whose context is too short for \name{}'s resident state (local window, anchors, summaries, and coarse entries) to fit in the GPU budget are excluded for every method. With $w=512$ and a mean span length of about 120 tokens, this state occupies roughly $2.6/|\bar{S}| + w/n$ of Full KV's memory for a context of $n$ tokens, so at a 10\% budget examples shorter than about 6.4K tokens are excluded.

\begin{table}[t]
\centering
\resizebox{\columnwidth}{!}{%
\begin{tabular}{lrrl}
\toprule
\textbf{Benchmark} & \textbf{Total} & \textbf{Kept} & \textbf{Context length} \\
\midrule
LongBench (16 tasks) & 3{,}750 & 3{,}200 & 6.4K--128K \\
RULER (13 tasks $\times$ 5 lengths) & 32{,}500 & 32{,}500 & 8K--128K \\ 
InfiniteBench (9 tasks) & 3{,}307 & 3{,}307 & up to 128K \\
NIAH (8 lengths $\times$ 11 depths) & 8{,}800 & 8{,}800 & 8K--128K \\
GSM8K (50-shot) & 1{,}319 & 1{,}319 & 12K--13K \\
\bottomrule
\end{tabular}
}
\caption{Evaluation sets. ``Kept'' denotes examples above
\name{}'s resident-state floor at a 10\% budget
(Qwen2.5-14B). The same subset is used for all methods.
GSM8K demonstrations are drawn from the training split,
which is not used to train \name{}.}
\label{tab:dataset_statistics}
\end{table}

\paragraph{LongBench}~\cite{bai-etal-2024-longbench} covers single- and multi-document QA, summarization, few-shot learning, synthetic retrieval, and code completion. We report the macro-average of the official per-task metrics (F1, ROUGE-L, accuracy, edit similarity).

\paragraph{RULER}~\cite{hsieh2024ruler} provides retrieval (single, multi-key, multi-value, multi-query needles), variable tracking, aggregation (common and frequent word extraction), and QA tasks at controlled lengths. We report the official string-match score averaged over 13 tasks and five lengths.

\paragraph{InfiniteBench}~\cite{zhang-etal-2024-bench} contains inputs that often exceed 100K tokens. We use En.QA, En.MC, En.Sum, En.Dia, Code.Debug, Math.Find, Retrieve.PassKey, Retrieve.Number, and Retrieve.KV with their official metrics, truncating inputs to 128K in the middle.

\paragraph{Needle-in-a-Haystack (NIAH)}~\cite{LLMTest_NeedleInAHaystack} inserts a single fact into Paul Graham essays at depths from 0\% to 100\% in steps of 10\%, at lengths of 8K, 16K, 24K, 32K, 48K, 64K, 96K, and 128K tokens, with 100 trials per cell. A trial counts as correct if the output contains the needle's answer.

\paragraph{GSM8K}~\cite{cobbe2021trainingverifierssolvemath} is used in a 50-shot setting (12K to 13K prompt tokens) to test whether compression preserves many in-context demonstrations. We report exact match after answer normalization.

\paragraph{SCBench}~\cite{li2025scbench} evaluates multi-turn reuse of a shared context. We use six tasks: Retr.KV, Retr.Prefix-Suffix, Retr.MultiHop, Code.RepoQA, En.QA, and Mix.RepoQA+KV, covering exact retrieval, semantic retrieval, and multi-task reasoning. We compress the context before the first query(query-agnostic) and report accuracy per turn.

\subsection{Baselines and Budget Matching}

All baselines use their official implementations where available. Each is configured so that its GPU-resident KV state equals the target budget; CPU memory is reported separately in Section~\ref{sec:efficiency_analysis}. Full KV is an uncompressed reference.

\paragraph{Eviction.} StreamingLLM~\cite{ICLR2024_5e5fd18f} keeps attention sinks and a recent window. H2O~\cite{zhang2023ho} keeps tokens with high accumulated attention. SnapKV~\cite{li2024snapkv} selects tokens from prompt-side attention in an observation window that includes the question. PyramidKV~\cite{cai2025pyramidkv} allocates the budget across layers in a pyramid. RocketKV~\cite{behnam2025rocketkv} evicts after prefill and applies sparse selection over the retained tokens during decoding. For all of them, the budget is the number of retained tokens.

\paragraph{Learned eviction.}
TRIM-KV~\cite{ICLR2026_c3887741} trains per-layer retention gates on a frozen backbone and evicts the lowest-scoring tokens when the cache exceeds the budget. For a fair comparison, we train its gates separately for each backbone using the same RedPajama~\cite{weber2024redpajama} corpus and training budget as \name{}, with no overlap with evaluation benchmarks.

\paragraph{Semantic compression.} ChunkKV~\cite{liu2026chunkkv} retains contiguous chunks, SemantiCache~\cite{wu2026semanticacheefficientkvcache} merges clusters of tokens into representative KV pairs, and SentenceKV~\cite{zhu2025sentencekv} keeps sentence-level vectors on GPU and retrieves token states from CPU during decoding. The budget is the number of retained units or active token states on GPU.

\paragraph{Low-rank compression.} OjaKV~\cite{zhu-etal-2026-ojakv} keeps high-error tokens at full rank and projects the rest onto an online low-rank basis. We adjust the projection rank for each backbone to match \name{}'s 10\% GPU-resident KV budget, ensuring a memory-matched comparison.

\paragraph{Recallable caches.} Quest~\cite{pmlr-v235-tang24l} keeps the full cache on GPU and attends to the top pages under query-aware bounds; it is matched on its attention-token budget and marked $\dagger$ in Table~\ref{tab:main_results}, since it does not reduce KV memory. InfLLM~\cite{10.5555/3737916.3741717} and RetroInfer~\cite{chen2025retroinfervectorstorageapproachscalable} offload the cache to CPU and retrieve blocks or tokens at each step; the budget covers their GPU-resident blocks, retrieved entries, and local window. ShadowKV~\cite{sun2025shadowkv} keeps low-rank pre-RoPE keys and landmarks on GPU, offloads values to CPU, and fetches a sparse token set per step; the budget covers its low-rank keys, landmarks, outliers, and sparse set.
\section{Extended Results}
\label{app:full_results}

This section reports the per-backbone results summarized in Table~\ref{tab:main_results}, together with \name{}'s accuracy relative to Full KV and its variance across router seeds.

\subsection{Per-Backbone Accuracy}

Table~\ref{tab:main_results_full} gives accuracy for every method, backbone, and benchmark at a 10\% GPU-resident KV budget. Three patterns hold across backbones. First, \name{} outperforms SentenceKV in all 20 settings, by 1.6 to 4.5 points, and TRIM-KV in 19 of 20. Second, \name{} is the best compressed method on LongBench for four of five backbones and on InfiniteBench for two (\textsc{Llama-3-8B} and \textsc{Mistral-7B}). Third, its deficit to ShadowKV is concentrated on NIAH, where it ranges from 2.0 to 6.8 points across backbones, and on RULER (1.6 to 4.5), while it is ahead of or tied with ShadowKV on LongBench for every backbone. On RULER and NIAH, the best compressed method is a recallable cache (ShadowKV or RetroInfer) for every backbone.

\begin{table*}[t]
\centering
\resizebox{\textwidth}{!}{%
\begin{tabular}{lcccccccccc}
\toprule
& \multicolumn{5}{c}{\textbf{LongBench}} & \multicolumn{5}{c}{\textbf{RULER}} \\
\cmidrule(lr){2-6}\cmidrule(lr){7-11}
\textbf{Method} & \rotatebox{55}{Llama-3.2-3B} & \rotatebox{55}{Llama-3-8B} & \rotatebox{55}{Llama-3.1-8B} & \rotatebox{55}{Mistral-7B} & \rotatebox{55}{Qwen2.5-14B} & \rotatebox{55}{Llama-3.2-3B} & \rotatebox{55}{Llama-3-8B} & \rotatebox{55}{Llama-3.1-8B} & \rotatebox{55}{Mistral-7B} & \rotatebox{55}{Qwen2.5-14B} \\
\midrule
Full KV & 41.76 & 45.26 & 50.42 & 45.83 & 54.68 & 74.63 & 66.38 & 87.91 & 69.64 & 89.28 \\
\midrule
\rowcolor{gray!12}\multicolumn{11}{l}{\emph{Eviction}} \\
StreamingLLM & 29.63 & 32.65 & 36.25 & 31.45 & 40.90 & 36.02 & 32.98 & 51.24 & 35.34 & 46.71 \\
H2O & 32.96 & 35.02 & 41.31 & 36.33 & 44.01 & 47.95 & 40.03 & 60.67 & 44.28 & 59.62 \\
SnapKV & 35.94 & 39.59 & 45.27 & 39.07 & 47.48 & 56.30 & 47.04 & 70.75 & 52.06 & 69.72 \\
PyramidKV & 35.52 & 38.78 & 45.07 & 38.83 & 48.66 & 58.83 & 49.42 & 71.48 & 50.40 & 71.18 \\
RocketKV & 36.16 & 40.01 & 45.47 & 39.49 & 49.91 & 63.18 & 52.62 & 78.66 & 56.66 & 78.03 \\
\rowcolor{gray!12}\multicolumn{11}{l}{\emph{Learned eviction}} \\
TRIM-KV & 35.72 & 39.21 & 45.86 & 39.11 & 49.73 & 58.29 & 49.21 & 76.09 & 53.52 & 73.59 \\
\rowcolor{gray!12}\multicolumn{11}{l}{\emph{Semantic compression}} \\
ChunkKV & 35.82 & 39.71 & 45.45 & 39.30 & 49.39 & 57.97 & 50.58 & 74.88 & 52.93 & 72.18 \\
SemantiCache & 35.78 & 39.68 & 45.17 & 39.91 & 49.24 & 57.59 & 49.17 & 73.32 & 52.27 & 73.40 \\
SentenceKV & 36.52 & 39.41 & 46.06 & 39.55 & 49.30 & 58.14 & 49.88 & 73.70 & 51.86 & 73.83 \\
\rowcolor{gray!12}\multicolumn{11}{l}{\emph{Low-rank compression}} \\
OjaKV & 30.96 & 33.49 & 38.25 & 34.61 & 43.40 & 50.73 & 42.03 & 64.18 & 43.57 & 62.83 \\
\rowcolor{gray!12}\multicolumn{11}{l}{\emph{Recallable caches}} \\
Quest$^\dagger$ & \underline{37.78} & 40.88 & 46.95 & \underline{41.47} & 50.86 & 62.40 & 53.30 & 79.92 & 56.90 & 77.64 \\
InfLLM & 36.16 & 39.58 & 45.23 & 40.00 & 49.61 & 53.81 & 48.05 & 72.09 & 50.96 & 67.93 \\
RetroInfer & 37.61 & \textbf{41.39} & 46.89 & 40.94 & 51.19 & \underline{66.28} & \underline{55.11} & \underline{81.26} & \underline{58.46} & \textbf{80.26} \\
ShadowKV & 37.45 & \underline{41.14} & \underline{47.28} & 40.98 & \underline{51.56} & \textbf{66.42} & \textbf{55.30} & \textbf{82.27} & \textbf{58.99} & \underline{79.83} \\
\midrule
\rowcolor{teal!8}\textbf{\name{} (ours)} & \textbf{38.42} & \underline{41.14} & \textbf{47.70} & \textbf{42.03} & \textbf{51.80} & 61.93 & 53.68 & 77.89 & 55.46 & 77.24 \\
\midrule
& \multicolumn{5}{c}{\textbf{InfiniteBench}} & \multicolumn{5}{c}{\textbf{NIAH}} \\
\cmidrule(lr){2-6}\cmidrule(lr){7-11}
\textbf{Method} & \rotatebox{55}{Llama-3.2-3B} & \rotatebox{55}{Llama-3-8B} & \rotatebox{55}{Llama-3.1-8B} & \rotatebox{55}{Mistral-7B} & \rotatebox{55}{Qwen2.5-14B} & \rotatebox{55}{Llama-3.2-3B} & \rotatebox{55}{Llama-3-8B} & \rotatebox{55}{Llama-3.1-8B} & \rotatebox{55}{Mistral-7B} & \rotatebox{55}{Qwen2.5-14B} \\
\midrule
Full KV & 32.68 & 28.31 & 45.13 & 31.76 & 48.71 & 85.72 & 80.46 & 97.11 & 83.65 & 98.22 \\
\midrule
\rowcolor{gray!12}\multicolumn{11}{l}{\emph{Eviction}} \\
StreamingLLM & 14.96 & 13.97 & 22.93 & 14.71 & 25.27 & 42.19 & 39.28 & 54.07 & 42.20 & 55.73 \\
H2O & 19.54 & 18.19 & 30.69 & 20.21 & 32.50 & 54.71 & 49.42 & 67.97 & 50.88 & 68.26 \\
SnapKV & 24.04 & 21.48 & 34.94 & 23.41 & 37.50 & 68.75 & 61.66 & 82.69 & 66.01 & 84.23 \\
PyramidKV & 23.13 & 21.37 & 35.03 & 22.81 & 38.50 & 71.57 & 61.58 & 85.92 & 66.21 & 84.51 \\
RocketKV & 25.79 & \underline{24.54} & 38.52 & 26.16 & 42.50 & 72.71 & 65.41 & 90.06 & 69.14 & 89.86 \\
\rowcolor{gray!12}\multicolumn{11}{l}{\emph{Learned eviction}} \\
TRIM-KV & 24.60 & 22.55 & 36.94 & 24.63 & 39.94 & 68.79 & 62.43 & 84.57 & 66.17 & 85.56 \\
\rowcolor{gray!12}\multicolumn{11}{l}{\emph{Semantic compression}} \\
ChunkKV & 24.71 & 22.45 & 36.52 & 24.15 & 40.31 & 68.43 & 60.97 & 83.52 & 63.81 & 81.21 \\
SemantiCache & 24.72 & 23.27 & 36.27 & 24.26 & 40.60 & 68.80 & 60.99 & 82.68 & 64.09 & 82.82 \\
SentenceKV & 24.32 & 23.28 & 36.12 & 24.63 & 40.60 & 70.16 & 61.57 & 82.54 & 65.20 & 81.53 \\
\rowcolor{gray!12}\multicolumn{11}{l}{\emph{Low-rank compression}} \\
OjaKV & 21.47 & 19.50 & 31.46 & 21.23 & 32.50 & 56.46 & 49.46 & 66.53 & 53.03 & 67.97 \\
\rowcolor{gray!12}\multicolumn{11}{l}{\emph{Recallable caches}} \\
Quest$^\dagger$ & \underline{26.45} & 24.24 & 39.30 & 26.05 & 42.48 & 76.11 & \underline{68.17} & 92.12 & 71.59 & 90.23 \\
InfLLM & 25.41 & 22.90 & 37.29 & 24.63 & 41.04 & 74.51 & 66.76 & 90.04 & 69.27 & 89.67 \\
RetroInfer & \textbf{27.31} & 24.16 & \textbf{39.92} & \underline{26.54} & \underline{42.65} & \textbf{76.56} & 67.74 & \textbf{92.86} & \textbf{72.31} & \textbf{91.75} \\
ShadowKV & 26.25 & 24.26 & \underline{39.49} & 26.51 & \textbf{43.72} & \underline{76.49} & \textbf{68.54} & \underline{92.16} & \underline{72.00} & \underline{91.23} \\
\midrule
\rowcolor{teal!8}\textbf{\name{} (ours)} & 26.13 & \textbf{25.12} & 38.47 & \textbf{26.76} & 42.50 & 74.53 & 66.05 & 85.36 & 67.93 & 85.12 \\
\bottomrule
\end{tabular}%
}
\caption{Per-backbone accuracy at a 10\% GPU-resident KV budget for all five backbones. Best compressed result in \textbf{bold}, second best \underline{underlined}. $^\dagger$Quest keeps the full KV cache on GPU and is matched on its attention-token budget. All results are means over three seeds.}
\label{tab:main_results_full}
\end{table*}

\subsection{Accuracy Relative to Full KV}

Table~\ref{tab:retention} reports \name{}'s accuracy as a percentage of Full KV. Retention is highest on LongBench (above 90\% for every backbone) and lowest on RULER on average, whose multi-key, multi-value, and variable-tracking tasks require several exact tokens at once. The two backbones extended to 128K with YaRN, \textsc{Llama-3-8B} and \textsc{Mistral-7B}, retain the least on RULER and NIAH, consistent with weaker long-range attention to recover from.

\begin{table}[t]
\centering
\small
\resizebox{\linewidth}{!}{%
\begin{tabular}{lccccc}
\toprule
\textbf{Backbone} & \textbf{LB} & \textbf{RULER} & \textbf{IB} & \textbf{NIAH} & \textbf{Avg} \\
\midrule
Llama-3.2-3B-Instruct & 92.0 & 83.0 & 80.0 & 86.9 & 85.5 \\
Llama-3-8B & 90.9 & 80.9 & 88.7 & 82.1 & 85.6 \\
Llama-3.1-8B-Instruct & 94.6 & 88.6 & 85.2 & 87.9 & 89.1 \\
Mistral-7B-Instruct-v0.3 & 91.7 & 79.6 & 84.3 & 81.2 & 84.2 \\
Qwen2.5-14B-Instruct & 94.7 & 86.5 & 87.3 & 86.7 & 88.8 \\
\midrule
Average & 92.8 & 83.7 & 85.1 & 85.0 & 86.6 \\
\bottomrule
\end{tabular}%
}
\caption{\name{} accuracy as a percentage of Full KV at a 10\% GPU-resident KV budget. LB: LongBench; IB: InfiniteBench.}
\label{tab:retention}
\end{table}

\subsection{Variance Across Router Seeds}
\label{app:seed_variance}

\name{}'s router is trained with three seeds per backbone, and all reported \name{} numbers are means over these seeds. Table~\ref{tab:seed_std} reports the standard deviations. They are at most 0.85 points and below 0.65 points in 18 of 20 settings, well below \name{}'s margins over SentenceKV (1.6--4.5 points). Its margins over ShadowKV on LongBench (0 to 1.1 points) are within about three standard deviations, so we do not claim that difference as significant.

\begin{table}[t]
\centering
\small
\resizebox{\linewidth}{!}{%
\begin{tabular}{lcccc}
\toprule
\textbf{Backbone} & \textbf{LB} & \textbf{RULER} & \textbf{IB} & \textbf{NIAH} \\
\midrule
Llama-3.2-3B-Instruct & 0.25 & 0.34 & 0.27 & 0.42 \\
Llama-3-8B & 0.31 & 0.81 & 0.25 & 0.31 \\
Llama-3.1-8B-Instruct & 0.37 & 0.44 & 0.64 & 0.40 \\
Mistral-7B-Instruct-v0.3 & 0.60 & 0.60 & 0.85 & 0.64 \\
Qwen2.5-14B-Instruct & 0.36 & 0.25 & 0.44 & 0.30 \\
\bottomrule
\end{tabular}%
}
\caption{Standard deviation of \name{} accuracy across three router seeds (points), 10\% GPU-resident KV budget.}
\label{tab:seed_std}
\end{table}
\section{Many-Shot and Multi-Turn Results}
\label{app:additional_results}

\paragraph{Many-shot GSM8K.}
Table~\ref{tab:e2} tests whether compression preserves many in-context demonstrations, using GSM8K with 50 demonstrations (12.3K to 13.1K prompt tokens) at a 10\% GPU-resident budget. Each demonstration is a short, self-contained example, so a method must keep enough of many of them rather than one piece of evidence. \name{} is the best method that fixes resolution at prefill on every backbone, ahead of SentenceKV by 0.9 to 1.7 points and of TRIM-KV by 1.0 to 2.1, and it retains 95.5\% of Full KV accuracy on average. It is within 0.6 points of ShadowKV on every backbone and ahead of it on \textsc{Llama-3-8B}, so the gap to exact recall seen on NIAH and RULER does not appear when the evidence is spread over many demonstrations.

\begin{table}[t]
\centering
\small
\setlength{\tabcolsep}{3pt}
\resizebox{\linewidth}{!}{%
\begin{tabular}{lcccccc}
\toprule
\textbf{Method} & \textbf{L3.2-3B} & \textbf{L3-8B} & \textbf{L3.1-8B} & \textbf{Mistral} & \textbf{Qwen} & \textbf{Avg.} \\
\midrule
Full KV & 73.24 & 68.46 & 84.76 & 56.41 & 93.48 & 75.27 \\
\midrule
SnapKV & 66.41 & 61.26 & 78.17 & 50.19 & 87.26 & 68.66 \\
PyramidKV & 67.10 & 60.73 & 79.23 & 49.81 & 88.17 & 69.01 \\
TRIM-KV & 68.23 & 62.40 & 80.14 & 51.33 & 89.46 & 70.31 \\
ChunkKV & 67.78 & 62.93 & 79.45 & 51.02 & 88.70 & 69.98 \\
SentenceKV & 68.54 & 63.15 & 80.21 & 51.71 & 89.16 & 70.55 \\
ShadowKV & \textbf{70.36} & \underline{64.29} & \textbf{81.65} & \textbf{53.22} & \textbf{91.43} & \textbf{72.19} \\
\rowcolor{teal!8}\name{} & \underline{69.83} & \textbf{64.52} & \underline{81.12} & \underline{52.92} & \underline{90.90} & \underline{71.86} \\
\bottomrule
\end{tabular}%
}
\caption{Exact-match accuracy on 50-shot GSM8K (12.3K--13.1K prompt tokens) at a 10\% GPU-resident KV budget. Best compressed result in \textbf{bold}, second \underline{underlined}.}
\label{tab:e2}
\end{table}

\paragraph{Multi-turn SCBench.}
Table~\ref{tab:e10} evaluates context reuse across turns on SCBench. The context is compressed once before the first query, so no method can use the queries to decide what to keep, and later turns ask about parts of the context that earlier turns did not need. We use six tasks, Retr.KV, Retr.Prefix-Suffix, Retr.MultiHop, Code.RepoQA, En.QA, and Mix.RepoQA+KV, and report accuracy averaged over tasks for each turn. Methods that discard entries degrade across turns: from turn 1 to turn 4, SnapKV loses about 13 points and TRIM-KV 9 to 10, whereas \name{}, SentenceKV, and ShadowKV, which keep every part of the context in some form, lose 2.1 to 2.8 points, against under 1 point for Full KV. By turn 4, \name{} is ahead of TRIM-KV by 9.0 to 9.4 points and of SnapKV by 14.4 to 14.9, and within 0.9 points of ShadowKV. \name{} has the smallest drop of all compressed methods on \textsc{Qwen2.5-14B} ($-2.2$) and the second smallest on \textsc{Llama-3.1-8B} ($-2.4$, against $-2.1$ for ShadowKV).

\begin{table}[t]
\centering
\small
\setlength{\tabcolsep}{2.5pt}
\resizebox{\linewidth}{!}{%
\begin{tabular}{lcccccccccc}
\toprule
& \multicolumn{5}{c}{\textbf{Qwen2.5-14B}} & \multicolumn{5}{c}{\textbf{Llama-3.1-8B}} \\
\cmidrule(lr){2-6}\cmidrule(lr){7-11}
\textbf{Method} & T1 & T2 & T3 & T4 & $\Delta$ & T1 & T2 & T3 & T4 & $\Delta$ \\
\midrule
Full KV & 82.74 & 82.36 & 82.91 & 81.83 & $-$0.91 & 76.42 & 76.81 & 75.93 & 75.67 & $-$0.75 \\
\midrule
SnapKV & 76.83 & 72.46 & 68.37 & 63.52 & $-$13.31 & 70.14 & 66.23 & 61.78 & 56.92 & $-$13.22 \\
TRIM-KV & 78.26 & 75.84 & 72.17 & 68.93 & $-$9.33 & 72.36 & 69.74 & 66.82 & 62.43 & $-$9.93 \\
SentenceKV & 79.43 & 78.62 & 77.83 & 76.94 & \underline{$-$2.49} & 73.58 & 72.93 & 72.14 & 70.82 & $-$2.76 \\
ShadowKV & \textbf{80.87} & \textbf{80.12} & \textbf{79.46} & \textbf{78.23} & $-$2.64 & \textbf{74.81} & \textbf{74.36} & \textbf{73.42} & \textbf{72.68} & \textbf{$-$2.13} \\
\rowcolor{teal!8}\name{} & \underline{80.14} & \underline{79.58} & \underline{78.86} & \underline{77.92} & \textbf{$-$2.22} & \underline{74.27} & \underline{73.81} & \underline{73.06} & \underline{71.85} & \underline{$-$2.42} \\
\bottomrule
\end{tabular}%
}
\caption{SCBench multi-turn accuracy per turn at a 10\% GPU-resident KV budget, with query-agnostic compression applied before the first turn and scores averaged over six tasks. $\Delta$: turn 4 minus turn 1, in points. Best compressed result in \textbf{bold}, second \underline{underlined}; for $\Delta$, the smallest drop.}
\label{tab:e10}
\end{table}
\section{Diagnosing the Exact-Retrieval Gap}
\label{app:niah_diagnosis}

This section analyzes where \name{} loses accuracy to ShadowKV, the strongest recallable cache, on NIAH and RULER. Tables~\ref{tab:niah_length_depth}--\ref{tab:ruler_tasks} use the main runs of Table~\ref{tab:main_results} on \textsc{Qwen2.5-14B-Instruct} and \textsc{Llama-3.1-8B-Instruct}, with per-example logging of routing, budget, and reconstruction decisions at the first answer token.

\begin{figure*}[t]
    \centering
    \includegraphics[width=\textwidth]{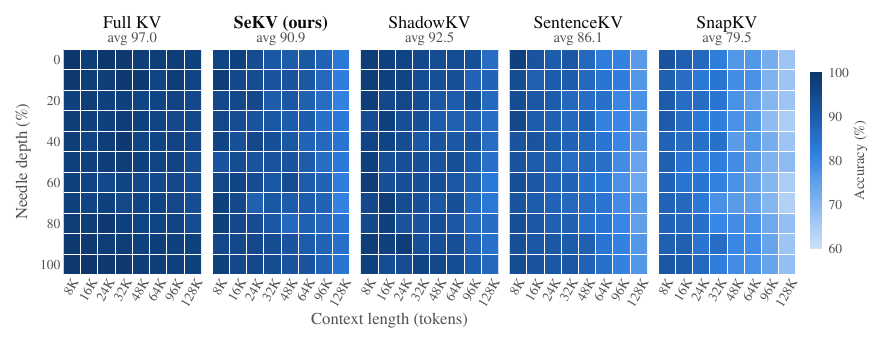}
    \caption{NIAH accuracy on \textsc{Llama-3.1-8B-Instruct} at a 10\% GPU-resident KV budget by context length and needle depth, 100 trials per cell, diagnostic setting. Averages over the grid are shown under each method.}
    \label{fig:niah_heatmap}
\end{figure*}

\paragraph{Diagnostic setting.} The ablations (Table~\ref{tab:ablations}), the budget sweep (Table~\ref{tab:budget}), router transfer and hyperparameter sensitivity (Appendix~\ref{app:hyper_sensitivity}), and Figures~\ref{fig:niah_length} and~\ref{fig:niah_heatmap} use a separate set of runs, which we call the \emph{diagnostic setting}; the span statistics and reconstruction analysis in Appendix~\ref{app:reconstruction} also use these runs. Absolute numbers in this setting differ from Table~\ref{tab:main_results}, so methods are compared only within it.

\paragraph{Accuracy by length and depth.}
Table~\ref{tab:niah_length_depth} breaks NIAH down by context length and needle depth. From 8K to 128K, \name{} loses 17.0 points on \textsc{Qwen2.5-14B} and 16.6 on \textsc{Llama-3.1-8B}, against 12.0 and 11.4 for ShadowKV and 3.4 and 5.6 for Full KV, so its gap to ShadowKV grows from 3.7 to 8.7 points and from 4.5 to 9.6 points. Depth matters much less: \name{}'s accuracy varies by 4.6 points across depth bins, against 17.0 across lengths. More spans compete for expansion at longer contexts, which points to routing and the budget cap rather than to where the needle sits. Figure~\ref{fig:niah_heatmap} shows the full grid on \textsc{Llama-3.1-8B} in the diagnostic setting, including SentenceKV and SnapKV, with the same pattern.

\begin{table*}[t]
\centering
\small
\setlength{\tabcolsep}{3.5pt}
\resizebox{\textwidth}{!}{%
\begin{tabular}{llcccccccccccc}
\toprule
& & \multicolumn{8}{c}{\textbf{Context length}} & \multicolumn{4}{c}{\textbf{Needle depth}} \\
\cmidrule(lr){3-10}\cmidrule(lr){11-14}
\textbf{Backbone} & \textbf{Method} & 8K & 16K & 24K & 32K & 48K & 64K & 96K & 128K & 0--20\% & 30--50\% & 60--80\% & 90--100\% \\
\midrule
\multirow{3}{*}{Qwen2.5-14B} & Full KV & 99.46 & 99.32 & 99.15 & 98.87 & 98.35 & 97.61 & 96.97 & 96.03 & 99.08 & 98.58 & 98.01 & 97.21 \\
 & ShadowKV & 96.46 & 95.20 & 93.87 & 92.87 & 91.25 & 89.04 & 86.67 & 84.48 & 92.66 & 91.84 & 90.78 & 89.64 \\
 & \cellcolor{teal!8}\name{} & \cellcolor{teal!8}92.77 & \cellcolor{teal!8}90.82 & \cellcolor{teal!8}89.04 & \cellcolor{teal!8}87.18 & \cellcolor{teal!8}84.47 & \cellcolor{teal!8}82.04 & \cellcolor{teal!8}78.88 & \cellcolor{teal!8}75.76 & \cellcolor{teal!8}87.47 & \cellcolor{teal!8}85.85 & \cellcolor{teal!8}84.25 & \cellcolor{teal!8}82.91 \\
\midrule
\multirow{3}{*}{Llama-3.1-8B} & Full KV & 99.23 & 98.94 & 98.59 & 98.00 & 97.16 & 96.29 & 95.00 & 93.67 & 97.97 & 97.47 & 96.90 & 96.10 \\
 & ShadowKV & 96.96 & 96.00 & 94.75 & 93.58 & 92.16 & 90.24 & 88.04 & 85.55 & 93.59 & 92.77 & 91.71 & 90.57 \\
 & \cellcolor{teal!8}\name{} & \cellcolor{teal!8}92.51 & \cellcolor{teal!8}91.02 & \cellcolor{teal!8}89.07 & \cellcolor{teal!8}87.56 & \cellcolor{teal!8}84.62 & \cellcolor{teal!8}82.35 & \cellcolor{teal!8}79.79 & \cellcolor{teal!8}75.96 & \cellcolor{teal!8}87.71 & \cellcolor{teal!8}86.09 & \cellcolor{teal!8}84.49 & \cellcolor{teal!8}83.15 \\
\bottomrule
\end{tabular}%
}
\caption{NIAH accuracy by context length (averaged over depths) and needle depth (averaged over lengths) at a 10\% GPU-resident KV budget, main runs. Both breakdowns average to the NIAH scores in Table~\ref{tab:main_results}.}
\label{tab:niah_length_depth}
\end{table*}

\paragraph{Where failures come from.}
Table~\ref{tab:niah_failures} classifies NIAH examples that Full KV answers correctly and \name{} does not. A failure is a \emph{routing miss} if no head expands the needle span, a \emph{budget cut} if the span passes its threshold but is dropped when the budget is filled, and a \emph{reconstruction failure} if the span is expanded and the answer is still wrong. All three contribute: reconstruction failures are the largest category (43\% and 42\%), followed by routing misses (38\% and 34\%) and budget cuts (19\% and 24\%). The learned rank already gives needle spans more components than the average span (20.8 against 17.3 for keys and 15.6 against 12.5 for values on \textsc{Qwen2.5-14B}), yet their value error remains higher (0.19 against 0.13), because a needle is an isolated fact that low-rank factors capture less well than the surrounding text. These rank and error statistics come from the main runs and are compared only with each other; Appendix~\ref{app:reconstruction} reports reconstruction error in the diagnostic setting. Needle tokens are about six times more likely to be anchors than tokens overall (5.7--6.3\% against about 1\%), so surprisal segmentation partly protects them. At the answer step, only 25--29\% of heads expand the needle span; the other heads attend to its coarse entry.

\begin{table}[t]
\centering
\small
\resizebox{\linewidth}{!}{%
\begin{tabular}{lcc}
\toprule
& \textbf{Qwen2.5-14B} & \textbf{Llama-3.1-8B} \\
\midrule
Failures analyzed (Full KV correct, \name{} wrong) & 126 & 108 \\
\quad Routing miss (\%) & 38.10 & 34.26 \\
\quad Budget cut (\%) & 19.05 & 24.07 \\
\quad Reconstruction failure (\%) & 42.85 & 41.67 \\
\midrule
Needle span: mean length (tokens) & 119.62 & 135.47 \\
Needle span: learned rank K / V & 20.84 / 15.63 & 19.41 / 14.72 \\
All spans: learned rank K / V & 17.29 / 12.48 & 16.31 / 11.95 \\
Needle span: V relative error & 0.19 & 0.22 \\
All spans: V relative error & 0.13 & 0.15 \\
Needle tokens that are anchors (\%) & 5.72 & 6.31 \\
Heads expanding the needle span (\%) & 28.64 & 25.19 \\
\bottomrule
\end{tabular}%
}
\caption{Sources of \name{}'s NIAH failures at the first answer token, main runs. The three failure categories sum to 100\%.}
\label{tab:niah_failures}
\end{table}

\paragraph{Oracle interventions.}
Table~\ref{tab:niah_oracle} changes one component at a time. \emph{Exact reconstruction} fetches expanded spans as exact KV instead of low-rank factors, \emph{oracle routing} forces the needle span to expand in every head, and \emph{no budget cap} removes the cap on expanded spans. These variants are diagnostic only and are not part of \name{}. No single change closes the gap. Exact reconstruction recovers 3.9 and 4.6 NIAH points and oracle routing 5.3 and 5.2, and together they reach 94.1 and 94.4, above ShadowKV (91.2 and 92.2) and within 4.1 and 2.7 points of Full KV. Removing the budget cap recovers 5.5 and 4.8 NIAH points and puts \name{} above ShadowKV on RULER, but raises GPU-resident KV to 14.8\% and 15.3\% of Full KV, half again the budget; this matches the narrowing gap at higher budgets in Table~\ref{tab:budget}. The gap to exact-recall caches therefore comes from both lossy values and missed routing, and closing it requires improving both, for example a higher rank or exact storage for values of high-surprisal spans and routing supervision at retrieval positions.

\begin{table}[t]
\centering
\small
\resizebox{\linewidth}{!}{%
\begin{tabular}{lcccc}
\toprule
& \multicolumn{2}{c}{\textbf{Qwen2.5-14B}} & \multicolumn{2}{c}{\textbf{Llama-3.1-8B}} \\
\cmidrule(lr){2-3}\cmidrule(lr){4-5}
\textbf{Variant} & NIAH & RULER & NIAH & RULER \\
\midrule
Full KV & 98.22 & 89.28 & 97.11 & 87.91 \\
ShadowKV & 91.23 & 79.83 & 92.16 & 82.27 \\
\midrule
\rowcolor{teal!8}\name{} & 85.12 & 77.24 & 85.36 & 77.89 \\
\quad + exact reconstruction & 89.04 & 79.05 & 89.96 & 80.15 \\
\quad + oracle routing & 90.37 & -- & 90.52 & -- \\
\quad + both & 94.10 & -- & 94.37 & -- \\
\quad + no budget cap$^\ast$ & 90.63 & 80.96 & 90.14 & 84.18 \\
\bottomrule
\end{tabular}%
}
\caption{Oracle interventions on \name{} at a 10\% GPU-resident KV budget, main runs. Oracle routing applies to NIAH only, where the needle position is known. $^\ast$Exceeds the budget: GPU-resident KV reaches 14.8\% (Qwen) and 15.3\% (Llama) of Full KV.}
\label{tab:niah_oracle}
\end{table}

\paragraph{RULER by task.}
Table~\ref{tab:ruler_tasks} breaks RULER down by task on \textsc{Qwen2.5-14B}. \name{} trails ShadowKV by 2.6 points on average, and 81\% of this gap comes from the multi-key, multi-value, and multi-query tasks (4.3 to 6.6 points each), which require several exact tokens to be recovered at once. On single-needle tasks the gap is 0.6 to 2.1 points, and on QA 1.2 to 2.3. \name{} is ahead on common and frequent word extraction (by 1.9 and 2.1 points), which aggregate information over the whole context and benefit from keeping every span in attention.

\begin{table*}[t]
\centering
\small
\resizebox{\textwidth}{!}{%
\begin{tabular}{lcccccccccccccc}
\toprule
\textbf{Method} & S-1 & S-2 & S-3 & MK-1 & MK-2 & MK-3 & MV & MQ & VT & CWE & FWE & QA-1 & QA-2 & \textbf{Avg.} \\
\midrule
Full KV & 98.48 & 96.40 & 94.46 & 96.29 & 93.54 & 91.18 & 91.62 & 89.70 & 84.05 & 78.97 & 77.08 & 85.63 & 83.24 & 89.28 \\
ShadowKV & 93.23 & 90.60 & 88.54 & 87.28 & 84.01 & 79.28 & 77.02 & 73.18 & 69.73 & 66.39 & 63.58 & 81.55 & 83.40 & 79.83 \\
\rowcolor{teal!8}\name{} & 92.59 & 89.69 & 86.46 & 82.99 & 78.94 & 73.81 & 70.41 & 67.49 & 66.36 & 68.25 & 65.67 & 79.28 & 82.18 & 77.24 \\
\bottomrule
\end{tabular}%
}
\caption{RULER accuracy per task on \textsc{Qwen2.5-14B-Instruct} at a 10\% GPU-resident KV budget, averaged over the five lengths, main runs. S: single needle; MK: multi-key; MV: multi-value; MQ: multi-query; VT: variable tracking; CWE/FWE: common/frequent word extraction.}
\label{tab:ruler_tasks}
\end{table*}
\section{Span Statistics and Reconstruction Quality}
\label{app:reconstruction}

\subsection{Span Statistics}

Table~\ref{tab:e6} reports how \name{} segments each benchmark and spends its GPU budget on \textsc{Qwen2.5-14B} and \textsc{Llama-3.1-8B} at a 10\% budget, in the diagnostic setting (Appendix~\ref{app:niah_diagnosis}). Anchors make up 0.9 to 1.2\% of tokens and spans average 113 to 131 tokens; longer contexts give slightly longer spans and fewer anchors (InfiniteBench: 129 to 131 tokens, 0.9\%). Values receive a higher learned rank than keys on every benchmark (13.8--15.3 against 10.8--12.2), consistent with values being harder to compress (Section~\ref{app:recon_quality}). How the budget is spent depends on context length. On LongBench, whose contexts are the shortest, the 512-token local window takes 58 to 61\% of the budget and reconstructed spans only 16 to 18\%; on InfiniteBench, the window takes 6\% and reconstructed spans 59 to 60\%. The resident state of anchors, summaries, and coarse entries takes 22 to 35\%. Zoom-in rates are lowest on LongBench and NIAH (3.1--3.7\%), higher on RULER (5.2--5.4\%), and highest on InfiniteBench (7.4--7.9\%).

\begin{table}[t]
\centering
\small
\setlength{\tabcolsep}{2.5pt}
\resizebox{\linewidth}{!}{%
\begin{tabular}{llccccccc}
\toprule
& & \textbf{Span} & \textbf{Anchor} & \textbf{Rank} & \multicolumn{3}{c}{\textbf{Budget share (\%)}} & \textbf{Zoom} \\
\cmidrule(lr){6-8}
\textbf{Backbone} & \textbf{Bench.} & (tok) & (\%) & K / V & Window & Resident & Recon. & (\%) \\
\midrule
\multirow{4}{*}{Qwen2.5-14B} & LB & 116.42 & 1.12 & 11.36 / 14.27 & 57.84 & 24.13 & 17.62 & 3.28 \\
 & RULER & 123.76 & 0.94 & 11.82 / 14.63 & 18.73 & 31.46 & 49.28 & 5.41 \\
 & IB & 131.24 & 0.87 & 12.17 / 15.34 & 5.72 & 35.18 & 58.63 & 7.86 \\
 & NIAH & 119.83 & 1.03 & 11.54 / 14.81 & 21.36 & 30.24 & 47.91 & 3.74 \\
\midrule
\multirow{4}{*}{Llama-3.1-8B} & LB & 112.68 & 1.18 & 10.84 / 13.76 & 61.27 & 22.38 & 15.92 & 3.11 \\
 & RULER & 121.35 & 0.98 & 11.28 / 14.19 & 20.46 & 29.82 & 49.17 & 5.16 \\
 & IB & 128.91 & 0.91 & 11.73 / 14.92 & 5.94 & 33.74 & 59.81 & 7.43 \\
 & NIAH & 117.46 & 1.07 & 11.12 / 14.38 & 23.18 & 28.64 & 47.67 & 3.59 \\
\bottomrule
\end{tabular}%
}
\caption{Span, budget, and zoom-in statistics at a 10\% GPU-resident KV budget. Span: mean span length; Anchor: share of tokens kept as anchors; Rank: mean learned rank of key and value factors. Budget share: fraction of the GPU-resident budget used by the local window, by the resident state (anchors, summaries, and coarse entries), and by reconstructed spans. Zoom: share of span, layer, and query-head decisions that expand the span. LB: LongBench; IB: InfiniteBench. Diagnostic setting.}
\label{tab:e6}
\end{table}

\subsection{Reconstruction Quality}
\label{app:recon_quality}

We measure the relative reconstruction error $\|X - \hat{X}\|_F / \|X\|_F$ of pre-RoPE keys and of values (Eq.~\ref{eq:reconstruction}), averaged over spans, and the share of spectral energy retained, on \textsc{Qwen2.5-14B-Instruct} at 128K context in the diagnostic setting (Table~\ref{tab:recon_rank}). Table~\ref{tab:recon_breakdown} breaks the error down by layer and span length, and Table~\ref{tab:recon_backbones} reports it for every backbone.

\paragraph{Values are harder to compress than keys.} At every rank, the value error is 1.5 to 2.3 times the key error, and rank 16 retains 94.7\% of the key energy but only 84.8\% of the value energy. The rank predictor reflects this and assigns values more components than keys on every backbone (13.2--15.0 against 10.4--11.6). This matches the diagnosis in Appendix~\ref{app:niah_diagnosis}, where exact reconstruction of expanded spans recovers 3.9 to 4.6 NIAH points.

\paragraph{Surprisal spans are easier to factorize.} At every rank, spans placed at surprisal boundaries have lower key and value error than fixed-length chunks of the same mean length (by 0.02 to 0.03), consistent with spans grouping tokens with related content.

\paragraph{The learned rank spends components where they matter.} With the learned rank, the average error (0.26 for keys and 0.43 for values) lies between that of fixed ranks 8 and 16, well above the error at rank 32 (0.14 and 0.27). Yet a fixed rank of 32 gives the same accuracy as the learned rank while using 1.6$\times$ the CPU memory (Table~\ref{tab:ablations}), so the average error over all spans is not what limits accuracy. What matters is the output of attention: on expanded spans, the relative error of the attention output is 0.09, much lower than the value error, because reconstructed tokens are combined with exact anchors, the local window, and coarse entries in one softmax.

\paragraph{Error grows with depth and span length.} The value error rises from 0.37 in the first third of the layers to 0.49 in the last, and from 0.22 for spans under 64 tokens to 0.52 for spans over 256 tokens, since longer spans hold more content per component. The pattern is similar across backbones: key errors range from 0.26 to 0.31 and value errors from 0.43 to 0.49, lowest on \textsc{Qwen2.5-14B} and highest on \textsc{Mistral-7B}.

\begin{table}[t]
\centering
\small
\setlength{\tabcolsep}{3.5pt}
\resizebox{\linewidth}{!}{%
\begin{tabular}{lcccccc}
\toprule
& \multicolumn{4}{c}{\textbf{Surprisal spans}} & \multicolumn{2}{c}{\textbf{Fixed chunks}} \\
\cmidrule(lr){2-5}\cmidrule(lr){6-7}
\textbf{Rank} & K err & K energy (\%) & V err & V energy (\%) & K err & V err \\
\midrule
8 & 0.41 & 83.19 & 0.61 & 62.79 & 0.44 & 0.64 \\
16 & 0.23 & 94.71 & 0.39 & 84.79 & 0.26 & 0.42 \\
32 & 0.14 & 98.04 & 0.27 & 92.71 & 0.17 & 0.30 \\
64 & 0.07 & 99.51 & 0.16 & 97.44 & 0.09 & 0.19 \\
\rowcolor{teal!8}Learned & 0.26 & 93.24 & 0.43 & 81.51 & 0.29 & 0.46 \\
\bottomrule
\end{tabular}%
}
\caption{Relative reconstruction error and retained spectral energy by rank on \textsc{Qwen2.5-14B-Instruct} at 128K context, averaged over spans. Keys are reconstructed before RoPE. Fixed chunks match the mean span length.}
\label{tab:recon_rank}
\end{table}

\begin{table}[t]
\centering
\small
\begin{tabular}{lcc}
\toprule
\textbf{Subset} & \textbf{K err} & \textbf{V err} \\
\midrule
Layers 1--16 & 0.24 & 0.37 \\
Layers 17--32 & 0.26 & 0.43 \\
Layers 33--48 & 0.28 & 0.49 \\
\midrule
Spans $<$64 tokens & 0.10 & 0.22 \\
Spans 64--256 tokens & 0.27 & 0.44 \\
Spans $>$256 tokens & 0.33 & 0.52 \\
\midrule
Attention output (expanded spans) & \multicolumn{2}{c}{0.09} \\
\bottomrule
\end{tabular}
\caption{Relative reconstruction error with the learned rank by layer and span length on \textsc{Qwen2.5-14B-Instruct} at 128K context. The last row is the relative error of the attention output when expanded spans are reconstructed instead of exact.}
\label{tab:recon_breakdown}
\end{table}

\begin{table}[t]
\centering
\small
\begin{tabular}{lcccc}
\toprule
\textbf{Backbone} & \textbf{K err} & \textbf{V err} & \textbf{Rank K} & \textbf{Rank V} \\
\midrule
Llama-3.2-3B & 0.29 & 0.46 & 10.38 & 13.21 \\
Llama-3-8B & 0.30 & 0.48 & 11.17 & 14.36 \\
Llama-3.1-8B & 0.27 & 0.44 & 11.12 & 14.38 \\
Mistral-7B & 0.31 & 0.49 & 11.64 & 14.97 \\
Qwen2.5-14B & 0.26 & 0.43 & 11.54 & 14.81 \\
\bottomrule
\end{tabular}
\caption{Relative reconstruction error with the learned rank, and mean learned rank, across backbones on NIAH at 128K context.}
\label{tab:recon_backbones}
\end{table}
\section{Efficiency Details}
\label{app:efficiency}

All measurements use batch size 1, 1{,}024 output tokens, greedy decoding, a 10\% GPU-resident KV budget, and a single A100 80GB GPU for every method, including Full KV. Throughput is (input $+$ output tokens) / total latency. GPU KV is the memory held by KV-related tensors on GPU, including reconstructed spans, and is at most 10\% of Full KV's KV memory for every compression method except Quest, which keeps the full cache on GPU. H2D is the host-to-device transfer per decoding step.

\begin{table*}[t]
\centering
\small
\resizebox{\textwidth}{!}{%
\begin{tabular}{llrrrrrrrr}
\toprule
\multicolumn{2}{l}{} & \multicolumn{4}{c}{\textbf{Qwen2.5-14B-Instruct}} & \multicolumn{4}{c}{\textbf{Llama-3.1-8B-Instruct}} \\
\cmidrule(lr){3-6}\cmidrule(lr){7-10}
Ctx & Method & Prefill (s) & Decode (ms/tok) & Peak GPU (GB) & CPU (GB) & Prefill (s) & Decode (ms/tok) & Peak GPU (GB) & CPU (GB) \\
\midrule
8K & Full KV & 1.64 & 30.40 & 34.12 & 0.00 & 0.94 & 20.12 & 20.15 & 0.00 \\
8K & Quest & 1.91 & 28.73 & 34.43 & 0.00 & 1.08 & 19.01 & 20.48 & 0.00 \\
8K & InfLLM & 1.82 & 39.17 & 31.87 & 1.51 & 1.14 & 29.37 & 18.64 & 1.04 \\
8K & SentenceKV & 1.79 & 36.48 & 31.92 & 1.48 & 1.16 & 24.61 & 18.88 & 1.03 \\
8K & ShadowKV & 1.94 & 25.83 & 32.14 & 0.82 & 1.19 & 16.93 & 19.02 & 0.55 \\
8K & \name{} & 1.86 & 33.30 & 32.36 & 0.70 & 1.22 & 22.67 & 19.13 & 0.48 \\
\midrule
32K & Full KV & 6.64 & 36.30 & 42.09 & 0.00 & 4.28 & 25.13 & 25.47 & 0.00 \\
32K & Quest & 7.02 & 25.93 & 44.37 & 0.00 & 4.69 & 18.37 & 26.07 & 0.00 \\
32K & InfLLM & 7.38 & 45.70 & 32.58 & 6.03 & 4.83 & 33.82 & 19.18 & 4.10 \\
32K & SentenceKV & 7.27 & 42.31 & 32.71 & 5.96 & 4.91 & 29.36 & 19.37 & 4.07 \\
32K & ShadowKV & 7.49 & 22.76 & 33.04 & 3.21 & 4.94 & 14.91 & 19.48 & 2.15 \\
32K & \name{} & 7.72 & 36.40 & 32.96 & 2.70 & 5.03 & 26.13 & 19.62 & 1.96 \\
\midrule
64K & Full KV & 16.18 & 45.10 & 51.14 & 0.00 & 9.91 & 31.32 & 32.64 & 0.00 \\
64K & Quest & 17.43 & 29.52 & 55.18 & 0.00 & 10.72 & 22.19 & 34.01 & 0.00 \\
64K & InfLLM & 18.23 & 56.31 & 33.72 & 12.09 & 11.18 & 40.54 & 20.27 & 8.18 \\
64K & SentenceKV & 17.96 & 48.20 & 33.84 & 12.01 & 11.02 & 34.89 & 20.13 & 8.04 \\
64K & ShadowKV & 18.41 & 24.91 & 34.19 & 6.43 & 11.39 & 18.66 & 20.39 & 4.27 \\
64K & \name{} & 18.62 & 43.80 & 33.96 & 5.40 & 11.47 & 30.82 & 20.63 & 3.94 \\
\midrule
128K & Full KV & 35.72 & 61.80 & 69.47 & 0.00 & 22.46 & 45.27 & 43.85 & 0.00 \\
128K & Quest & 38.16 & 37.89 & 71.23 & 0.00 & 23.19 & 29.38 & 46.38 & 0.00 \\
128K & InfLLM & 40.87 & 78.37 & 35.06 & 24.16 & 25.12 & 56.72 & 21.04 & 16.38 \\
128K & SentenceKV & 40.46 & 64.71 & 35.22 & 23.58 & 24.89 & 46.83 & 21.21 & 16.19 \\
128K & ShadowKV & 41.19 & 31.68 & 35.42 & 12.84 & 25.46 & 24.71 & 21.57 & 8.82 \\
128K & \name{} & 39.37 & 56.30 & 35.31 & 10.70 & 25.73 & 40.56 & 21.48 & 7.84 \\
\bottomrule
\end{tabular}%
}
\caption{Prefill time, decode latency, peak GPU memory, and CPU memory from 8K to 128K context.}
\label{tab:e3a}
\end{table*}

\begin{table}[t]
\centering
\small
\resizebox{\linewidth}{!}{%
\begin{tabular}{lrrrr}
\toprule
Ctx & \multicolumn{2}{c}{Qwen2.5-14B} & \multicolumn{2}{c}{Llama-3.1-8B} \\
\cmidrule(lr){2-3}\cmidrule(lr){4-5}
 & GPU KV (GB) & H2D (MB) & GPU KV (GB) & H2D (MB) \\
\midrule
8K & 0.14 & 12.0 & 0.10 & 9.9 \\
32K & 0.51 & 22.7 & 0.39 & 18.4 \\
64K & 1.08 & 40.1 & 0.76 & 33.2 \\
128K & 2.17 & 68.2 & 1.53 & 57.3 \\
\bottomrule
\end{tabular}%
}
\caption{\name{}'s GPU-resident KV memory and host-to-device transfer per decoding step.}
\label{tab:sekv_gpukv}
\end{table}

\paragraph{Where \name{}'s time goes.}
Table~\ref{tab:e3b} breaks down \name{}'s prefill and decode time on \textsc{Qwen2.5-14B}. Surprisal and SVD add 13--15\% to prefill. During decoding, attention and the rest of the model take 25--29 ms per token regardless of context length, while routing, reconstruction, and the transfer time not hidden behind computation grow with the number of spans. Reuse of spans expanded at the previous step rises from 72\% at 8K to 89\% at 128K, which keeps per-step transfer small as context grows. The effective host-to-device bandwidth is 3--5 GB/s (e.g., 68.2 MB in 16.9 ms at 128K), far below PCIe limits, which suggests that the transfer path, not its volume, is the main source of \name{}'s latency gap to ShadowKV.

\begin{table}[t]
\centering
\small
\resizebox{\linewidth}{!}{%
\begin{tabular}{lrrrr}
\toprule
Component (Qwen2.5-14B) & 8K & 32K & 64K & 128K \\
\midrule
Prefill: transformer (s) & 1.64 & 6.72 & 16.20 & 34.40 \\
Prefill: surprisal, chunked LM head (s) & 0.10 & 0.39 & 0.87 & 1.73 \\
Prefill: segmentation and SVD (s) & 0.12 & 0.61 & 1.55 & 3.24 \\
\midrule
Decode: routing (ms/tok) & 2.80 & 4.00 & 5.30 & 8.20 \\
Decode: H2D, total (ms/tok) & 2.60 & 7.40 & 11.30 & 16.90 \\
Decode: H2D, not overlapped (ms/tok) & 1.20 & 3.60 & 5.90 & 9.60 \\
Decode: reconstruction (ms/tok) & 1.90 & 3.30 & 5.80 & 9.40 \\
Decode: attention and rest (ms/tok) & 27.40 & 25.50 & 26.80 & 29.10 \\
Spans reused from previous step (\%) & 72 & 78 & 83 & 89 \\
\bottomrule
\end{tabular}%
}
\caption{Breakdown of \name{}'s prefill and decode time. Decode rows that sum to the total use the non-overlapped H2D time.}
\label{tab:e3b}
\end{table}

\paragraph{Short contexts.}
Table~\ref{tab:e13} reports end-to-end latency at 4K and 8K input. At these lengths the KV cache is small relative to the weights, so compression saves little: \name{} is 9--13\% slower than Full KV, faster than SentenceKV, and slower than ShadowKV and StreamingLLM, and peak GPU memory differs by about 3 GB or less across methods.

\begin{table}[t]
\centering
\small
\resizebox{\linewidth}{!}{%
\begin{tabular}{llrrr}
\toprule
In / Out & Method & Total (s) & Thr. (tok/s) & Peak GPU (GB) \\
\midrule
4096 / 1024 & Full KV & 31.22 & 164.00 & 32.41 \\
 & StreamingLLM & 25.37 & 201.81 & 30.76 \\
 & ChunkKV & 33.46 & 153.02 & 31.24 \\
 & SentenceKV & 37.18 & 137.71 & 31.37 \\
 & ShadowKV & 27.91 & 183.45 & 31.68 \\
 & \name{} & 34.13 & 150.01 & 31.83 \\
\midrule
4096 / 4096 & Full KV & 125.36 & 65.35 & 33.12 \\
 & StreamingLLM & 99.83 & 82.06 & 31.08 \\
 & ChunkKV & 136.21 & 60.14 & 31.74 \\
 & SentenceKV & 151.79 & 53.97 & 31.86 \\
 & ShadowKV & 110.42 & 74.19 & 32.17 \\
 & \name{} & 140.51 & 58.30 & 32.28 \\
\midrule
8192 / 1024 & Full KV & 32.77 & 281.23 & 34.12 \\
 & StreamingLLM & 26.98 & 341.59 & 31.23 \\
 & ChunkKV & 35.13 & 262.34 & 31.76 \\
 & SentenceKV & 39.15 & 235.40 & 31.92 \\
 & ShadowKV & 28.39 & 324.62 & 32.14 \\
 & \name{} & 35.96 & 256.28 & 32.36 \\
\midrule
8192 / 4096 & Full KV & 134.73 & 91.20 & 34.82 \\
 & StreamingLLM & 107.43 & 114.38 & 31.72 \\
 & ChunkKV & 141.37 & 86.92 & 32.17 \\
 & SentenceKV & 164.82 & 74.55 & 32.36 \\
 & ShadowKV & 118.97 & 103.29 & 32.58 \\
 & \name{} & 151.64 & 81.03 & 32.74 \\
\bottomrule
\end{tabular}%
}
\caption{Short-context runtime on \textsc{Qwen2.5-14B-Instruct}, batch size 1, one A100 80GB.}
\label{tab:e13}
\end{table}
\subsection{Cost Model}
\label{app:cost_model}

Let $L$ be the number of layers, $H_q$ and $H_{kv}$ the numbers of query and KV heads, $d_h$ the head dimension, $n$ the context length, $a$ the number of anchors, $w$ the local window, $m$ the number of spans, and $b$ the bytes per scalar (2 for bf16, 4 for the fp32 factors on CPU). Full KV stores $2LH_{kv}d_h n$ scalars.

\paragraph{GPU.} Before any span is expanded, \name{} keeps exact KV for anchors and the local window, one coarse entry per span and KV head, and one $d'$-dimensional routing summary per span and query head:
\begin{equation}
\begin{aligned}
M_{\text{GPU}} ={}& \underbrace{2LH_{kv}d_h(a+w)}_{\text{anchors, window}} \\
&+ \underbrace{2LH_{kv}d_h\,m}_{\text{coarse entries}} + \underbrace{LH_q d'\,m}_{\text{summaries}}
\end{aligned}
\end{equation}
scalars. Relative to Full KV, this is about $a/n + w/n + (1 + H_q d'/2H_{kv}d_h)/|\bar{S}|$, which defines the minimum budget used to filter examples in Appendix~\ref{app:benchmark_baseline_details}. Reconstructed spans fill the rest of the budget.

\paragraph{CPU.} Each span stores rank-$r_S$ factors of its keys and values for every layer and KV head:
\begin{equation}
M_{\text{CPU}} = \sum_{S} 2LH_{kv}\, r_S\, (|S| + d_h)
\end{equation}
scalars, ignoring the $r_S$ singular values. Because $r_S \ll d_h$ for most spans, this is a small fraction of Full KV, and the learned rank reduces it further than a fixed rank (Table~\ref{tab:ablations}).

\paragraph{Transfer.} At step $t$, only factors of spans that are expanded and were not expanded at step $t-1$ are fetched, so the transfer is $\sum_{S \in \mathcal{Z}_t \setminus \mathcal{Z}_{t-1}} r_S(|S| + d_h)$ scalars per key or value, layer, and KV head, where $\mathcal{Z}_t$ is the set of expanded spans. It depends on the zoom-in rate and on reuse, not on $n$ directly (Table~\ref{tab:sekv_gpukv}).

\paragraph{Compute.} Routing costs $O(LH_q m d')$ per token, linear in the number of spans rather than tokens. Reconstructing an expanded span costs $O(|S|\, r_S\, d_h)$ per head, and attention runs over $a + w + m$ entries plus the tokens of expanded spans.
\section{Router Cost, Transfer, and Sensitivity}
\label{app:hyper_sensitivity}

\paragraph{Training cost.}
Table~\ref{tab:e9b} lists the trainable parameters and training time of the router for each backbone. The router adds 2.8M to 7.9M parameters, 0.054\% to 0.086\% of the backbone, and trains in 1.8 to 5.2 hours on one $8\times$A100 (80GB) node (14 to 42 GPU-hours). It is trained once per backbone and reused for every benchmark and budget.

\begin{table}[t]
\centering
\small
\setlength{\tabcolsep}{3.5pt}
\resizebox{\linewidth}{!}{%
\begin{tabular}{lcccc}
\toprule
\textbf{Backbone} & \textbf{Params} & \textbf{\% of model} & \textbf{GPU-hours} & \textbf{Wall-clock (h)} \\
\midrule
Llama-3.2-3B & 2.8M & 0.086 & 14.4 & 1.8 \\
Llama-3-8B & 4.3M & 0.054 & 24.8 & 3.1 \\
Llama-3.1-8B & 4.3M & 0.054 & 27.2 & 3.4 \\
Mistral-7B & 4.3M & 0.058 & 26.4 & 3.3 \\
Qwen2.5-14B & 7.9M & 0.054 & 41.6 & 5.2 \\
\bottomrule
\end{tabular}%
}
\caption{Router training cost per backbone on one $8\times$A100 (80GB) node.}
\label{tab:e9b}
\end{table}

\paragraph{Router transfer.}
Table~\ref{tab:e9a} tests how much the router depends on its training backbone and corpus. Applying the router trained on \textsc{Llama-3-8B} to \textsc{Llama-3.1-8B}, which has the same architecture, costs 1.8 points on NIAH and 2.1 on RULER relative to a router trained on \textsc{Llama-3.1-8B} itself, about half of the 3.4 and 3.8 points lost by training-free routing (Table~\ref{tab:ablations}). A router can therefore be reused within a model family, but training on the target backbone, which takes a few hours, recovers the rest. The training corpus matters little: routers for \textsc{Qwen2.5-14B} trained on LongAlpaca or PG-19 with the same tokens and steps stay within 1.2 points of the RedPajama router on both benchmarks, and no corpus is best on both, so \name{}'s results do not depend on the choice of RedPajama.

\begin{table}[t]
\centering
\small
\resizebox{\linewidth}{!}{%
\begin{tabular}{llcc}
\toprule
\textbf{Router trained on} & \textbf{Evaluated on} & \textbf{NIAH} & \textbf{RULER} \\
\midrule
Llama-3-8B & Llama-3-8B & 66.05 & 53.68 \\
\rowcolor{teal!8}Llama-3.1-8B & Llama-3.1-8B & 90.88 & 78.94 \\
Llama-3-8B & Llama-3.1-8B & 89.13 & 76.82 \\
\midrule
\rowcolor{teal!8}RedPajama (default) & Qwen2.5-14B & 91.62 & 79.49 \\
LongAlpaca & Qwen2.5-14B & 91.17 & 78.72 \\
PG-19 & Qwen2.5-14B & 90.46 & 80.14 \\
\bottomrule
\end{tabular}%
}
\caption{Router transfer across backbones (top) and training corpora (bottom) at a 10\% GPU-resident KV budget, diagnostic setting (Appendix~\ref{app:niah_diagnosis}). Corpus comparisons use matched training tokens and steps. Default routers are shaded.}
\label{tab:e9a}
\end{table}

\paragraph{Hyperparameter sensitivity.}
Table~\ref{tab:e11} varies one hyperparameter at a time on \textsc{Qwen2.5-14B-Instruct} at a 10\% budget, keeping the others at their defaults. For every hyperparameter, the default is within 0.5 points of the best value on all four benchmarks. Three settings matter. $R_{\max}$ and $\beta$ trade accuracy for CPU memory: $R_{\max}=8$ halves CPU memory (5.5 GB) at a cost of 1.7 to 3.3 points, while $R_{\max}=64$ adds under 0.3 points because the learned rank barely grows (13.2 to 13.7); $\beta=0.01$ gains under 0.5 points for 20\% more CPU memory, and $\beta=1.0$ saves 27\% at a cost of 0.8 to 1.7 points. The surprisal threshold $\alpha$ has an optimum near 1.0: lower values create too many short spans and higher values too few anchors, costing up to 2.9 points at $\alpha=2.0$. A longer local window helps slightly ($w=1024$: under 0.4 points), and $L_{\min}$ has no measurable effect (within 0.4 points).

\begin{table}[t]
\centering
\small
\setlength{\tabcolsep}{3pt}
\resizebox{\linewidth}{!}{%
\begin{tabular}{llcccccc}
\toprule
& & & & & & \textbf{Mean} & \textbf{CPU} \\
\textbf{Param.} & \textbf{Value} & \textbf{LB} & \textbf{RULER} & \textbf{IB} & \textbf{NIAH} & \textbf{rank} & \textbf{(GB)} \\
\midrule
\multirow{4}{*}{$R_{\max}$} & 8 & 49.37 & 76.62 & 40.81 & 88.36 & 7.12 & 5.47 \\
 & 16 & 50.86 & 78.31 & 41.93 & 90.43 & 10.24 & 8.14 \\
 & \cellcolor{teal!8}32 & \cellcolor{teal!8}51.80 & \cellcolor{teal!8}79.49 & \cellcolor{teal!8}42.50 & \cellcolor{teal!8}91.62 & \cellcolor{teal!8}13.18 & \cellcolor{teal!8}10.70 \\
 & 64 & 51.97 & 79.76 & 42.68 & 91.91 & 13.74 & 11.19 \\
\midrule
\multirow{4}{*}{$\alpha$} & 0.5 & 50.71 & 77.62 & 41.48 & 89.83 & 12.87 & 10.43 \\
 & \cellcolor{teal!8}1.0 & \cellcolor{teal!8}51.80 & \cellcolor{teal!8}79.49 & \cellcolor{teal!8}42.50 & \cellcolor{teal!8}91.62 & \cellcolor{teal!8}13.18 & \cellcolor{teal!8}10.70 \\
 & 1.5 & 51.56 & 78.98 & 42.17 & 91.13 & 13.42 & 10.91 \\
 & 2.0 & 50.43 & 76.94 & 41.32 & 88.76 & 13.65 & 11.11 \\
\midrule
\multirow{3}{*}{$w$} & 256 & 50.98 & 78.13 & 41.82 & 90.26 & 13.18 & 10.70 \\
 & \cellcolor{teal!8}512 & \cellcolor{teal!8}51.80 & \cellcolor{teal!8}79.49 & \cellcolor{teal!8}42.50 & \cellcolor{teal!8}91.62 & \cellcolor{teal!8}13.18 & \cellcolor{teal!8}10.70 \\
 & 1024 & 52.03 & 79.84 & 42.71 & 91.93 & 13.18 & 10.70 \\
\midrule
\multirow{3}{*}{$L_{\min}$} & 8 & 51.51 & 79.12 & 42.24 & 91.31 & 13.26 & 10.77 \\
 & \cellcolor{teal!8}16 & \cellcolor{teal!8}51.80 & \cellcolor{teal!8}79.49 & \cellcolor{teal!8}42.50 & \cellcolor{teal!8}91.62 & \cellcolor{teal!8}13.18 & \cellcolor{teal!8}10.70 \\
 & 32 & 51.62 & 79.27 & 42.31 & 91.38 & 13.09 & 10.62 \\
\midrule
\multirow{3}{*}{$\beta$} & 0.01 & 52.12 & 79.93 & 42.79 & 92.08 & 15.64 & 12.84 \\
 & \cellcolor{teal!8}0.1 & \cellcolor{teal!8}51.80 & \cellcolor{teal!8}79.49 & \cellcolor{teal!8}42.50 & \cellcolor{teal!8}91.62 & \cellcolor{teal!8}13.18 & \cellcolor{teal!8}10.70 \\
 & 1.0 & 50.96 & 77.83 & 41.72 & 89.94 & 9.87 & 7.83 \\
\bottomrule
\end{tabular}%
}
\caption{Hyperparameter sensitivity on \textsc{Qwen2.5-14B-Instruct} at a 10\% GPU-resident KV budget, diagnostic setting (Appendix~\ref{app:niah_diagnosis}). One hyperparameter is varied at a time; defaults are shaded. Mean rank averages the learned key and value ranks; CPU memory at 128K context.}
\label{tab:e11}
\end{table}
\section{Additional Zoom-In Analysis}
\label{app:additional_zoom_in}

\begin{figure*}[t]
    \centering
    \includegraphics[width=\textwidth]{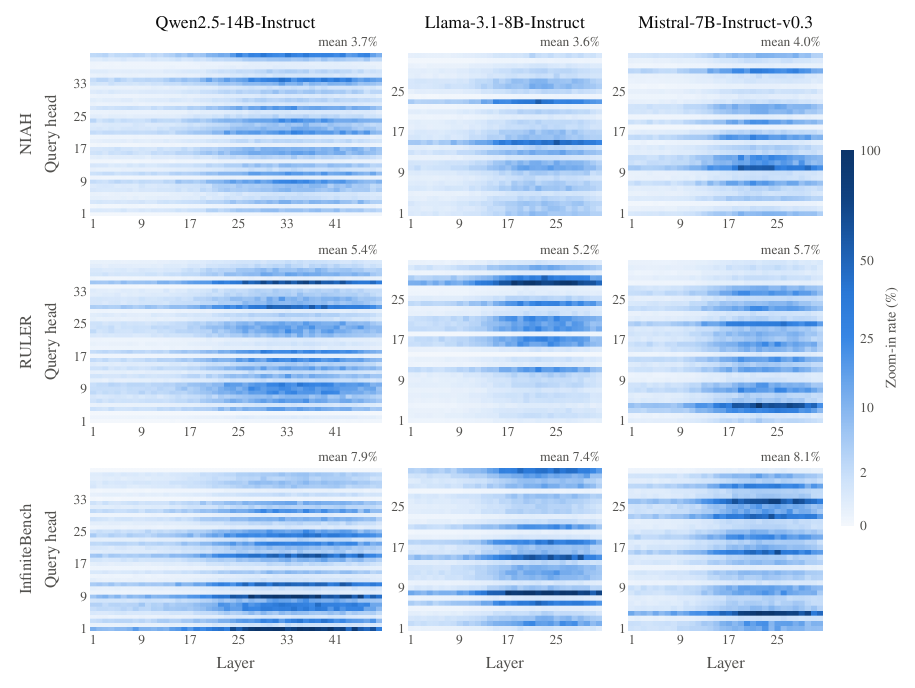}
    \caption{Zoom-in rate per layer and query head for three backbones (columns) and three benchmarks (rows) at a 10\% GPU-resident KV budget. The mean rate of each panel is shown above it. The color scale is square-root compressed and shared across panels.}
    \label{fig:app_zoom_heatmaps}
\end{figure*}

Figure~\ref{fig:app_zoom_heatmaps} extends Figure~\ref{fig:zoom_heatmap} to three backbones and three benchmarks at a 10\% GPU-resident budget. The patterns of Section~\ref{sec:zoom_analysis} hold in every panel. Mean zoom-in rates follow the same order on every backbone, NIAH (3.6--4.0\%) below RULER (5.2--5.7\%) below InfiniteBench (7.4--8.1\%). Layers in the first third of each network expand fewer than 2\% of spans on average, and the 10\% most active heads account for 49--63\% of expansions. Per-head rates for the same backbone correlate only weakly across benchmarks (Spearman 0.07--0.49), and the number of heads that expand more than half of all spans depends on the task: 27--61 heads on InfiniteBench and 13--18 on RULER, but at most 6 on NIAH. These heads behave like retrieval heads, but which heads take this role changes with the task, so a single global threshold or a fixed set of retrieval heads chosen offline would over-expand for some tasks and under-expand for others. This aligns with the global-threshold cost in Table~\ref{tab:ablations}.

\end{document}